\pdfoutput=1
\documentclass[11pt]{article}
\usepackage{arabtex}
\usepackage[]{EMNLP2023}
\usepackage{xspace,mfirstuc,tabulary}
\usepackage{times}
\usepackage{latexsym}
\usepackage{lipsum}

\usepackage[export]{adjustbox}

\usepackage{amsmath}
\usepackage{hyperref}
\usepackage{cleveref}

\usepackage{rotating}
\usepackage{tablefootnote}
\usepackage{adjustbox}
\usepackage{booktabs}
\usepackage{multirow}

\usepackage{microtype}
\usepackage{xspace}
\usepackage{mathtools}
\usepackage{amssymb}
\usepackage{footnote}
\usepackage{color,soul}
\usepackage{graphicx}
\usepackage{xstring}
\usepackage{comment}
\usepackage{adjustbox}
\usepackage{float}
\restylefloat{table}
\usepackage{array,booktabs,makecell}

\newcolumntype{H}{>{\setbox0=\hbox\bgroup}c<{\egroup}@{}}
\usepackage{caption}
\usepackage{subcaption}

\usepackage[T1]{fontenc}
\usepackage[utf8]{inputenc}
\usepackage{xstring}
\usepackage{cleveref}
 \usepackage{arabtex}
\usepackage{utf8}
\setcode{utf8}
\usepackage{color}
\usepackage{arydshln}
\usepackage{adjustbox}
\usepackage{float}
\restylefloat{table}
\usepackage{array,booktabs,makecell}
\usepackage{multirow}
\usepackage{graphicx}
\usepackage{color, colortbl}
\usepackage{xstring}
\usepackage{arydshln}
\usepackage{appendix}
\usepackage{microtype}

\usepackage{inconsolata}

\usepackage{arydshln}
\newcolumntype{H}{>{\setbox0=\hbox\bgroup}c<{\egroup}@{}}
\usepackage{cancel}
\usepackage{ulem,xpatch}

\newcommand\blfootnote[1]{%
  \begingroup
  \renewcommand\thefootnote{}\footnote{#1}%
  \addtocounter{footnote}{-1}%
  \endgroup
}

\newcommand{\dolphin}{Dolphin}

\title{\includegraphics[scale=0.078]{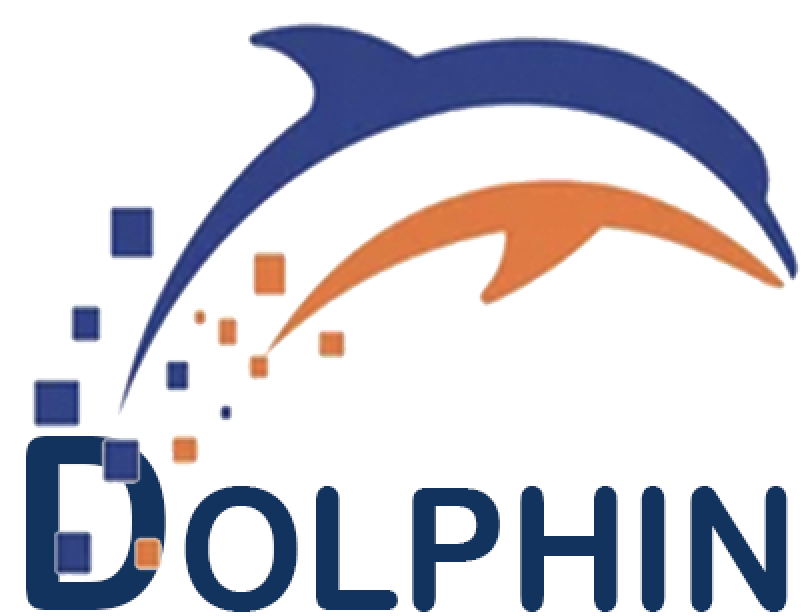}: A Challenging and Diverse Benchmark for Arabic NLG}

\author{\normalsize El Moatez Billah Nagoudi$^{\xi,\star}$ ~~~~~~~~~~~~~~~~~~~~ AbdelRahim Elmadany$^{\xi,\star}$ \\ \normalsize \textbf{Ahmed Oumar El-Shangiti}$^{\lambda}$~~~~~~~~~~~~~~~~~~~\textbf{Muhammad Abdul-Mageed}$^{\xi,\lambda,\star}$ \\
\normalsize $^{\xi}$ Deep Learning \& Natural Language Processing Group,
  The University of British Columbia\\\normalsize  $^{\lambda}$Department of Natural Language Processing \& Department of Machine Learning, MBZUAI\\ %
  \texttt{\normalsize \{moatez.nagoudi,a.elmadany,muhammad.mageed\}@ubc.ca}}

\setcode{utf8}
\setarab 
\begin{document}

\maketitle

%We hope~\dolphine will foster advancements in Arabic NLG by encouraging the development of more robust, context-aware, and linguistically sophisticated NLG models.

% \begin{abstract}
\section*{~~~~~~~~~~~~~~~~~~~~~~~~~~~~~Abstract}
We present~\textit{\dolphin}, a novel benchmark that addresses the need for a natural language generation (NLG) evaluation framework dedicated to the wide collection of Arabic languages and varieties. The proposed benchmark encompasses a broad range of $13$ different NLG tasks, including dialogue generation, question answering, machine translation, summarization, among others. ~\textit{\dolphin} comprises a substantial corpus of $40$ diverse and representative public datasets across $50$ test splits, carefully curated to reflect real-world scenarios and the linguistic richness of Arabic. It sets a new standard for evaluating the performance and generalization capabilities of Arabic and multilingual models, promising to enable researchers to push the boundaries of current methodologies. We provide an extensive analysis of~\dolphin, highlighting its diversity and identifying gaps in current Arabic NLG research. We also offer a public leaderboard that is both interactive and modular and evaluate several models on our benchmark, allowing us to set strong baselines against which researchers can compare.\footnote{\href{https://dolphin.dlnlp.ai/}{https://dolphin.dlnlp.ai/}.} 
% \end{abstract}
\section{Introduction}\label{intro}
%%%%%%%%%%%%%%%%%%%%%%%%%
 ~\blfootnote{ $^{\star}${Equal contributions.}}
Natural language generation (NLG) systems attempt to produce coherent, contextually appropriate, and linguistically accurate human-like language. These systems have a wide range of applications in everyday life, including in recreation, education, health, etc. The recent rise of generative models has transformed these NLG systems, making them more relevant and engaging than before. 
Crucial to measuring the performance of NLG systems are high-quality benchmarks. In particular, they provide standardized frameworks for comparing and quantitatively assessing different algorithms, models, and techniques. For NLG, benchmarks define specific criteria and metrics for evaluating performance, allowing for objectively gauging the strengths and limitations of different approaches and encouraging healthy competition. NLG benchmarks can also facilitate reproducibility and promote transparency across different studies, acting as a catalyst for advancement in the field. 

 \begin{figure}[t]

 \includegraphics[scale=0.14]{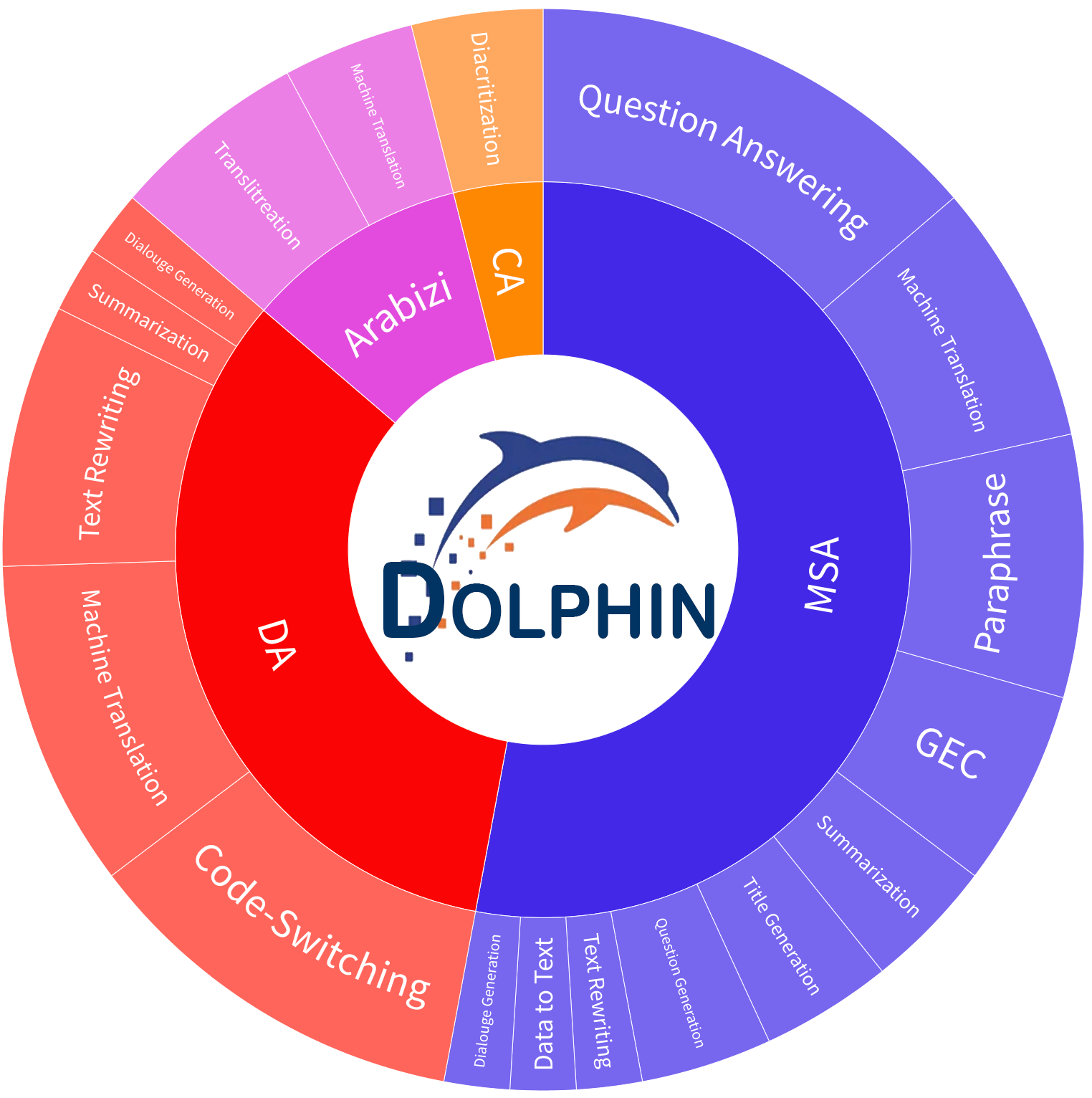}
  \caption{\textit{\dolphin}~task clusters and taxonomy.~\textbf{GEC}:
grammatical error correction. \textbf{CA}: Classical Arabic. \textbf{DA}:~Dialectal Arabic. \textbf{MSA}: Modern Standard Arabic.}  \label{fig:sarlu_tax} 
 \end{figure}

Despite of this significance, efforts for developing nuanced NLG benchmarks that can allow us to track and guide performance on particular languages remain limited. For Arabic, a wide collection of languages and diverse varieties, there is currently no sizeable benchmark that caters to the needs of the community. In this work, we present a large benchmark for Arabic, dubbed\textit{~\dolphin}, to bridge this gap. Our novel benchmark is carefully curated to represent real-world usage of Arabic at scale.~\dolphin~covers \textit{Classical Arabic (CA)}, a pre-modern standardized form of Arabic used for old poetry and religious discourse that continues to be employed for literary expression and oration, \textit{Modern Standard Arabic (MSA)}, a modern descendent of CA used in formal settings and in pan-Arab media, \textit{dialectal Arabic (DA)}, such as varieties used in everyday communication in the different Arab countries.~\dolphin~also encompasses text written in both Arabic and Latin scripts, the latter usually referred to as \textit{Arabizi}. The benchmark is comprised of $13$ different generation tasks based on $40$ different datasets across $50$ test splits, making it by far the largest Arabic NLG benchmark to date and among the largest for any group of languages. 

We build~\dolphin~on top of exclusively \textit{public} datasets, adding a number of newly developed datasets of our creation. This makes~\dolphin~accessible and easy to use. Our benchmark is accompanied by a modular leaderboard with a unified evaluation metric, i.e., a\textit{~\dolphin~score}. The leaderboard is designed to serve as a central hub for tracking and showcasing the performance of NLG systems. It functions as a dynamic and transparent platform where users can submit their models to compare their results against the state-of-the-art approaches. It also encourages a culture of transparency and detailed model description. 

Overall, we make the following contributions: \textbf{(1)} We introduce a novel benchmark for Arabic NLG that is large, public, diverse, and inclusive. \textbf{(2)} We develop a dynamic leaderboard with a rich array of best design principles to facilitate the measurement of progress in the field. \textbf{(3)} We evaluate a wide host of Arabic and multilingual models on our benchmark, offering strong baselines. \textbf{(4)} We analyze our benchmark to identify gaps in existing work, hoping to help guide future directions. The rest of the paper is organized as follows: In Section~\ref{sec:literature_light}, we provide an overview of related work. Section~\ref{sec:dolphin} introduces~\dolphin~design principles and task clusters. In Section~\ref{sec:Eval}, we present evaluations of the pretrained models on~\dolphin, and discuss the results we acquire. We conclude in Section~\ref{sec:conc}.

% In Section~\ref{sec:llms}, we describe the Arabic and multilingual sequence-to-sequenc pretrained language models.

% \mam{ tasks: 13
% datasets: 37
% test splits: 50
% MSA: \hl{28}
% DA: \hl{17}
% Arabizi: \hl{5}}

%%%%%%%%%%%%%%%%%%%%%%

\section{Related Works}\label{sec:literature_light}
\begin{figure}[t]
 \includegraphics[scale=0.32,left]{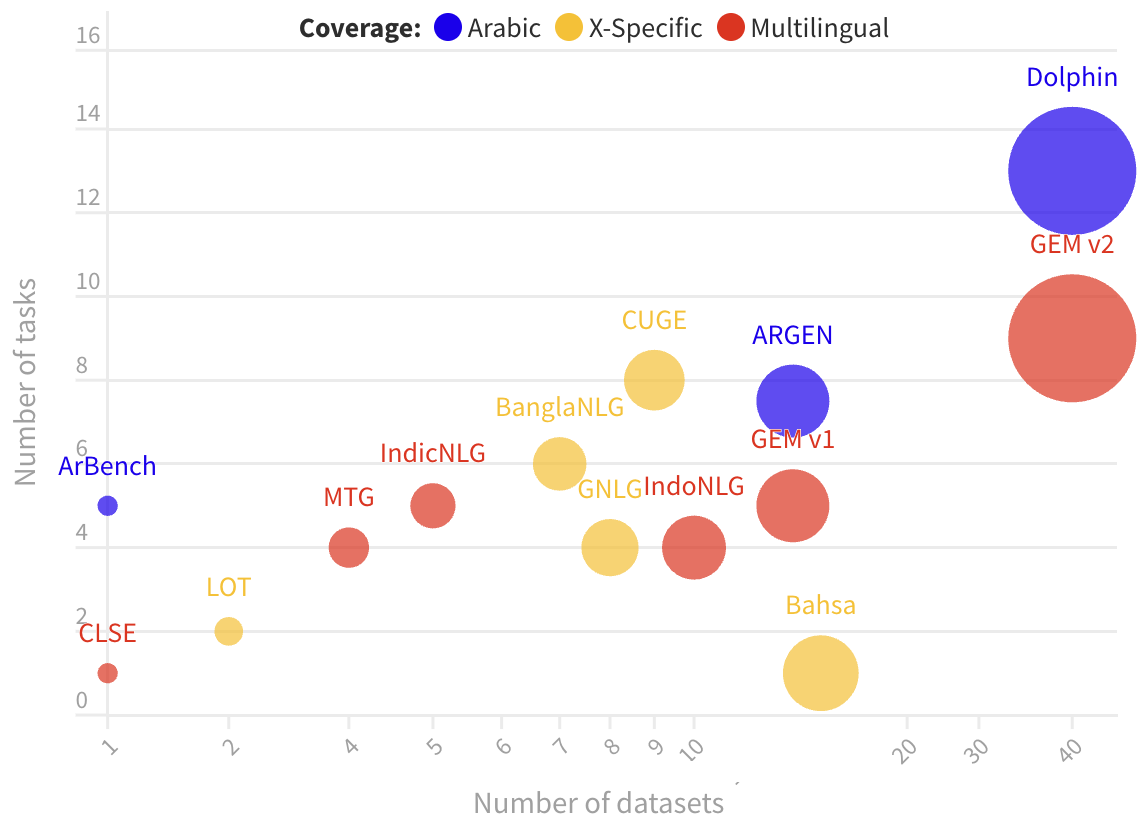}
  \caption{Comparison of the number of datasets and tasks supported by the Arabic (including \textit{\dolphin}), X-specific, and Multilingual NLG benchmarks. } \label{fig:bench_comp} 
 \end{figure}

Existing NLG benchmarks can be classified into three distinct categories: \textit{Arabic-specific}, \textit{X-specific} (where X refers to languages other than Arabic, such as English, Chinese, etc.), and \textit{multilingual} benchmarks. In this section, we provide a brief overview of each category, highlighting their respective characteristics and scope. We offer more details on target languages, dataset sizes, and the breadth of tasks ~\dolphin~covers in Appendix~\ref{app:RW_app}. Table~\ref{tab:bench_comp} and Figure~\ref{fig:bench_comp} offer a summary of comparisons between~\dolphin~and other benchmarks.

% \subsection{ Benchmarks}
% \begin{itemize}
%     \item \noindent\textbf{GLGE}\cite{liu2020glge} The General Language Generation Evaluation (GLGE) benchmark is a multi-task benchmark for assessing NLG in the English language generalization abilities. It has eight English language generation tasks, including dialogue, generative question answering, question generation, and text summarization.
% \end{itemize}
% \textit{data-to-text}, \textit{dialog response generation}, \textit{generative question answering}, \textit{question generation},   \textit{simplification}, and \textit{summarization}.

% propose SuperGLUE, a benchmark styled after GLUE with a new set of more challenging tasks. SuperGLUE is built around eight tasks and  arranged into four task clusters: QA, NLI, WSD, and coreference resolution. The benchmark is accompanied by a leaderboard with a single-number performance metric (i.e., the \textit{SuperGLUE score}). 
%%%%%%%%
 \noindent \textbf{Arabic Benchmarks.}
%%%%%%%%
\newcite{sajjad2020arabench} introduce AraBench, a machine translation (MT) evaluation benchmark consisting of five datasets for dialectal Arabic to English translation. AraOpus-20~\cite{nagoudi-2022-turjuman} is  another MT benchmark of parallel sentences between Arabic and $20$
languages.~\newcite{nagoudi2022_arat5} introduce ArGen, an Arabic NLG benchmark composed of $19$ datasets covering seven tasks. In comparison,~\dolphin~is much larger, composed of exclusively \textit{public} datasets, and covers more varieties. It is also the only benchmark accompanied by a leaderboard.

%%%%%%%%
\noindent \textbf{X-Specific Benchmarks.}
%%%%%%%%
\noindent\newcite{liu2020glge} propose GLGE, a generation benchmark for English covering eight datasets across four tasks.  CUGE \cite{yao2021cuge} and LOT ~\cite{10.1162/tacl_a_00469} are two Chinese benchmarks that cover both language understanding and generation tasks. BanglaNLG~\cite{bhattacharjee2022banglanlg} is a generation benchmark designed for Bangala comprising seven datasets across six tasks. ~\newcite{guntara-etal-2020-benchmarking} and \newcite{doan-etal-2021-phomt} present two MT benchmarks for Bahasa Indonesia and Vietnamese languages, respectively.  

%\hl{https://arxiv.org/abs/2112.13610}

%%%%%%%%%%%%%%%%%%%%%%%%%%%%%%
%%%%%%%%%%%%%%%%%%%%%%%%%%%
%%%%%%%%%%%%%%%%%%%%%%%%%%%
\begin{table*}[]
\centering
 \renewcommand{\arraystretch}{1}
% \resizebox{\columnwidth}{!}{%
\resizebox{0.95\textwidth}{!}{%

\begin{tabular}{cllp{0.45\linewidth}ccc}
\toprule
\textbf{Category} & \textbf{Benchmark} &  \textbf{Reference}  & \textbf{Task Clusters}  & \textbf{Lang} & \textbf{Datasets} & \textbf{\#Clusters} \\
\toprule
\multirow{5}{*}{\rotatebox[origin=]{90}{\textbf{\colorbox{green!10}{\small{Arabic}}}}}

% &\multirow{2}{*}{\textbf{\textbf{Dolphin}}} 

&  \multirow{2}{*}{\includegraphics[scale=0.04]{Image/Dolphin_Logo.png}} &  \multirow{2}{*}{\textbf{\textcolor{blue}{\textit{Our work}}}} & \textit{ADT,  CS, DRG, DT, GES, MT,  NTG, PPH,  } &   \multirow{2}{*}{Ar}       &       \multirow{2}{*}{$40$}     &     \multirow{2}{*}{$13$ }      \\
& &   & \textit{QA, QG, TRW, TRS, TS} &       &           &         \\

\cdashline{2-7} 
&{ArBench}& \newcite{sajjad2020arabench}  & \textit{MT} & Ar     &        $5$     &     $1$       \\
&{AraOPUS-20}& \newcite{nagoudi-2022-turjuman}  & \textit{MT} & Ar     &        $1$     &     1       \\
&{ARGEN}   & \newcite{nagoudi2022_arat5}  &  \textit{CS,  MT, NTG,  PPH, QG, TS,  TRS} & Ar     &     $13$         &     $7$        \\ \midrule

\multirow{5}{*}{\rotatebox[origin=]{90}{\textbf{\colorbox{blue!5}{\small{X-Specific}}}}}
&{GNLG}    & \newcite{liu2020glge} & \textit{DRG, DT, TT, TS} & En      &       $8$     &    $4$       \\  

&{BanglaNLG}    & \newcite{bhattacharjee2022banglanlg}&  \textit{MT, TS, QA, DRG, NTG, CLTS}  & Bn      &    $7$         &    $6$       \\  
&{CUGE}    &  \newcite{yao2021cuge}  & \textit{QA, DR, TS, DT, DRG, MT, CLTS, MC}   & Zh      &        $9$    &   $8$        \\

&{Bahasa Indonesia}    &  \newcite{guntara-etal-2020-benchmarking}  & \textit{MT}  & Id      &        $14$    &   $1$      \\

&{PhoMT}    &  \newcite{doan-etal-2021-phomt}  & \textit{MT}  & Vi      &        $1$    &   $1$      \\

&{LOT}    &  \newcite{10.1162/tacl_a_00469}  & \textit{RES, DS}  & Zh      &        $2$    &   $2$      \\ \midrule

\multirow{6}{*}{\rotatebox[origin=]{90}{\textbf{\colorbox{red!10}{\small{Multilingual}}}}}

&{CLSE}     &  \newcite{chuklin2022clse}& DRG  &   $3$    &    $1$          &      $1$ \\

&{GEM\textsubscript{v1}}     &  \newcite{gehrmann2021gem}&  \textit{DRG, DT, RES, TS,  SMP } &  $18$    &    $13$          &      $5$      \\

&{GEM\textsubscript{v2}}   &  \newcite{gehrmann2022gemv2}  & \textit{DRG, DT,  PPH, QA, QG, RES, SLG, SMP, TS} & $51$     &          $40$      &      $9$     \\

&{IndicNLG}   &  \newcite{kumar-etal-2022-indicnlg}  & \textit{NTG, TS, PPH, QG, BG} & $11$     &          $5$      &      $5$     \\

&{MTG}   &  \newcite{chen-etal-2022-mtg}  & \textit{SG, QG, NTG, TS} & $5$     &          $4$      &      $4$     \\

&{IndoNLG}    & \newcite{cahyawijaya2021indonlg}& \textit{TS, QA, CC, MT}  &  $3$      &     $10$  &   $4$         \\  
\toprule
\end{tabular}}

\caption{Comparison of  NLG benchmarks proposed in the literature across the different covered task clusters.  \textbf{ADT}: Arabic text diacritization. \textbf{CS}: Code-Switching. \textbf{DRG}: dialogue response generation. \textbf{DT}: data-to-text.    \textbf{GEC}: grammatical error correction. \textbf{MT}: machine translation. \textbf{NTG}: news title generation. \textbf{PPH}:  paraphrase.  \textbf{QA}: question answering.  \textbf{QG}: question generation. \textbf{RES}: reasoning.  \textbf{SLG}: slide generation. \textbf{SMP}: text simplification. \textbf{TRS}: transliteration. \textbf{TRW}: text rewriting. \textbf{TS}: text summarization. \textbf{TT}: table to text. \textbf{CLTS}: cross-lingual text summarization. \textbf{MC}: math computation. \textbf{DR}: document retrieval. \textbf{DS}: discourse structure. \textbf{CC}: chit-chat. \textbf{BG}: biography generation. \textbf{SG}: story generation.}  
\label{tab:bench_comp}

\end{table*}

\noindent \textbf{Multi-Lingual NLG Benchmarks.} The generation evaluation and metrics benchmark (\textit{GEM\textsubscript{v1}})~\cite{gehrmann2021gem} is a multilingual benchmark environment for NLG.  GEM\textsubscript{v1} features $18$ languages across $13$ datasets spanning five tasks. ~\newcite{gehrmann2022gemv2} propose a second version, GEM\textsubscript{v2}, with a new set of datasets and more challenging tasks. This new version supports $40$ documented datasets in $51$ languages. Other multilingual NLG benchmarks include  CLSE \cite{chuklin2022clse}, IndoNLG~\cite{cahyawijaya2021indonlg}, IndicNLG~\cite{kumar-etal-2022-indicnlg}, and MTG~\cite{chen-etal-2022-mtg}. 
As Figure~\ref{fig:bench_comp} shows, compared to these benchmarks,~\dolphin~is the largest both in terms of the number of tasks and datasets. We now introduce~\dolphin.

%\noindent\textbf{GLUECoS.}

%\hl{https://aclanthology.org/2022.emnlp-main.360/}

% \input{Tables/dolphin_tasks}
% %%%%%%%%%%%%%%%%%%%%%%

\section{Dolphin Benchmark}\label{sec:dolphin}
%\lipsum[1]
%We present \textit{Dolphin}, a comprehensive, challenging, diverse, and unified Arabic NLG evaluation benchmark. 
Our objective is to provide a comprehensive and challenging benchmark for natural language generation that enables the assessment of language models and the tracking of progress in Arabic. To attain this objective, we develop~\dolphin~, considering several design principles that we will now elucidate.%. We  elaborate on these principles in the following section.

\subsection{Design Principles}\label{subsec:design_prin}
\noindent\textbf{Wide, diverse coverage.} As our goal is to offer a demanding and diverse benchmark, we incorporate \textit{as many datasets from as many tasks as is feasible}. This allows comprehensive evaluations of LMs. It also facilitates painting as complete a picture as possible of the limits of current methods across the different tasks. Reflecting this principle, our benchmark is large. It comprises $40$ distinct datasets, covering $13$ different task clusters.

\noindent\textbf{Public datasets.} 
A major principle in choosing our datasets is public accessibility as it enables researchers to train and evaluate models without incurring expenses associated with acquiring private data. For this reason, all our $40$ datasets are publicly available. 

\noindent\textbf{Rich linguistic variability.} In order to accurately reflect the multifaceted and diverse nature of Arabic languages and dialectal varieties, we strategically incorporate datasets collated from an array of sources, each corresponding to different sociological and orthographic traditions. Specifically, we construct ~\dolphin~considering four major variants of Arabic: Arabizi (an unconventional method where Arabic is transcribed in Latin script); Classical Arabic (CA); Dialectal Arabic (DA) from a myriad of cities, countries, and regions; and Modern Standard Arabic (MSA). The heterogeneous nature of our datasets allows for a comprehensive representation of Arabic across wide linguistic nuances and orthographic traditions. Refer to Figure~\ref{fig:sarlu_tax} for an illustrative depiction of the distribution of our datasets over various Arabic varieties for each specific task. Table~\ref{tab:dolphin-stats_vara} provides a quantitative description of these varieties in~\dolphin.

\noindent\textbf{Standard evaluation metrics.} Most generation tasks can be evaluated using traditional automated metrics such as BLEU~\cite{papineni2002bleu} and ROUGE~\cite{lin2004rouge}. Both of these metrics evaluate the n-gram overlap between a reference text and the generated text. Nevertheless, in many tasks (e.g., question generation, open domain generation, title generation) there are multiple valid ways to produce a given text. In our benchmark, in addition to F\textsubscript{1}, BLEU, and ROUGE, we use several other evaluation metrics such MaxMatch (M2)~\cite{dahlmeier2012better} for grammatical error correction, and Character Error Rate (CER)~\cite{morris2004} for diacritization.

\noindent\textbf{Modular, interactive leaderboard.}
To support future research, we develop a public leaderboard that enables the evaluation of multilingual and Arabic LMs on~\dolphin. Our leaderboard is interactive and provides detailed metadata about the corpora such as size, training, development, and test splits, data sources (e.g., URL, GitHub), and citations to publications. The leaderboard also offers details of language models assessed such as the number of parameters, epochs to conversion, pretraining and finetuning information, etc. We provide a screenshot from our leaderboard in Figure~\ref{fig:leaderboard}. We now introduce each of the task clusters in~\dolphin.

\begin{table}[H]
\centering
 \renewcommand{\arraystretch}{1}
\resizebox{0.9\columnwidth}{!}{%
% \resizebox{0.95\textwidth}{!}{%
\begin{tabular}{l|c|c|c}
\toprule
 \textbf{Task Variety}  & \textbf{\# Clusters} & \textbf{\# Datasets} & \textbf{\# Test Sets} \\
\toprule
     \multirow{1}{*}{\textit{Arabizi $\rightarrow$ X}}  &  $1$ &  $2$  & $2$   \\
     \multirow{1}{*}{\textit{Arabizi $\rightarrow$ MSA}}  &  $1$ &  $3$  & $3$   \\ \midrule
    %%%%%%%%%% CA %%%%%%%%%%%
     \multirow{1}{*}{\textit{CA $\rightarrow$ CA}}  &  $1$ &  $1$  & $1$    \\ \midrule
    %%%%%%%%%% DA %%%%%%%%%%%
    \multirow{1}{*}{\textit{DA $\rightarrow$ DA}}  &  $2$ &  $2$  & $3$   \\
    \multirow{1}{*}{\textit{DA $\rightarrow$ MSA}}  &  $1$ &  $1$  & $4$   \\ 
    \multirow{1}{*}{\textit{DA $\rightarrow$ En}}  &  $1$ &  $1$  & $5$   \\ 
    \multirow{1}{*}{\textit{DA-X $\rightarrow$ X}}  &  $1$ &  $1$  & $6$   \\ \midrule
    %%%%%%%%%% Table %%%%%%%%%%% 
    \multirow{1}{*}{\textit{Table $\rightarrow$ MSA}}  &  $1$ &  $1$  & $1$   \\ \midrule
     %%%%%%%%%% MSA %%%%%%%%%%% 
    \multirow{1}{*}{\textit{MSA $\rightarrow$ MSA}}  &  $7$ &  $21$  & $21$   \\\midrule
     %%%%%%%%%% X %%%%%%%%%%% 
    \multirow{1}{*}{\textit{X $\rightarrow$ MSA}}  &  $1$ &  $2$  & $4$   \\

  \toprule
  
\end{tabular}
}
\caption{Descriptive statistics of the linguistic diversity in~\dolphin~across the different data splits.}  
\label{tab:dolphin-stats_vara}
\end{table}

\subsection{Task Clusters}\label{subsec:clusters}
%%%%%%%%%%%%%%%%%%%%%%%%%%%%%%%%%%%%%%%%%%%%%%%%%
\dolphin~involves $50$ test sets curated from $40$ datasets. We arrange~\dolphin~into 13 task clusters, as follows: (1) machine translation, (2) code-switching, (3) text summarisation, (4) news title generation, (5) question answering, (6) question generation, (7) transliteration, (8) paraphrasing, (9) text rewriting, (10) diacritization, (11) data-to-text, (12) dialogue generation, and (13) grammatical error correction. Appendix Table~\ref{tab:dolphin-stats} shows a summary of the data splits across datasets and task clusters in~\dolphin.   We present each task cluster in~\dolphin~next.

\subsubsection{Machine Translation}\label{sec:MT}

The MT cluster is built around three tasks:  
\noindent\textbf{(1)~\textit{X~$\rightarrow$~MSA}.} In this task,  we test the ability of the models to translate from six foreign languages into MSA. We use the UN parallel corpus~\cite{ziemski2016united}, a dataset covering the six official UN languages (i.e., Arabic, Chinese, English, French, Russian, and Spanish). The UN corpus consists of development and test sets only.\footnote{$4$K sentences that are aligned across all official languages.} For training, we randomly select $50$K \textit{X}-Arabic parallel sentences from the multilingual corpus MultiUN~\cite{eisele2010multiun} where \textit{X} is a language from the six official languages.

\noindent\textbf{(2) Arabizi~$\rightarrow$~X.}  The goal of this task is to translate from Arabizi dialectal text\footnote{ \textit{\textbf{Arabizi}}  is the romanization of Arabic script~\cite{darwish2013arabizi}. In this task, we investigate the Algerian and Moroccan Arabizi.} into one of two foreign languages  French and English.  For this, we use Darija~\cite{outchakoucht2021moroccan} and NArabizi~\cite{seddah-etal-2020-building}.

\noindent\textbf{(3) Dialects $\rightarrow$ English.} For this task, we focus on MT from six Arabic dialects into English using the MDP corpus~\cite{bouamor2014multidialectal}. MDP is a human-translated collection of $1$K sentences in Egyptian, Tunisian, Jordanian, Palestinian, and Syrian Arabic, in addition to English. For training, we use the $10$K  MSA-English  manually translated sentences proposed by~\newcite{bouamor2018madar} under a `zero-shot' condition.\footnote{Due to lexical overlap between Arabic dialects and MSA, this is not zero-shot in the strict sense of the word.}

%\cite{bouamor2014multidialectal} contains parallel corpus of MSA, EGY, PA, SY, TN, JO and ENG. We map MSA and its dialects to ENG. We split the data to 70/10/20 for \textit{train, dev and test} respectively.\\
%From SADID Corpus \cite{abid-2020-sadid} we built two datasets. First dataset is translation from Egyptian and Levantine to English, the dataset is splitted into \textit{dev} and \textit{devtest} with 3K sentences each.
%and the other being from English, Egyptian and Levantine to MSA(more of this in \ref{subsec:X->Arabic}).
%This is because the original splits do not cover the 4-way translation in all splits(for instance dev and devtests splits contain only lev, egy and eng translation with MSA missed)

% The authors \cite{meftouh-etal-2015-machine}

\subsubsection{Code-Switching}\label{subsec:CST}
The purpose of the code-switching (CS) task cluster is to translate Arabic dialect text that includes code-switching with a foreign language into that foreign language. For this, we create six new human-written (natural) code-switched parallel test datasets, under two tasks: 
\noindent\textbf{\textit{(1)~DIA-FR~$\rightarrow$~FR.}} This consists of $300$ code-switched Arabic-French tweets collected from Algerian, Moroccan, and Tunisian Twitter. \noindent\textbf{\textit{(2)~DIA-EN~$\rightarrow$~EN.}}~This is collected from Egyptian, Jordanian, and Palestinian Twitter and consists of $300$ code-switched Arabic-English posts. 
\noindent For both of these DIA-FR and DIA-EN tasks, a human translation is performed by one native speaker from each dialect with semi-native English/French fluency. For these two tasks, we perform experiments under the zero-shot setting. That is, we use no actual \textit{code-switched training} data. Rather, we extract $50$K MSA-English and MSA-French sentences from AraOPUS-20~\cite{nagoudi-2022-turjuman} that we use for \textit{monolingual training}. We then extract $50$ pairs from each code-switched dialect pair for development and test on the $250$ remainder sentences.

\subsubsection{Text Summarization}

For the text summarization (TS)  cluster, we use the following five Arabic and multilingual (including Arabic) publicly available datasets: (1)~MassiveSum~\cite{varab-schluter-2021-massivesumm},  (2)~XLSum~\newcite{hasan2021xlsum}, (3)~CrossSum~\cite{bhattacharjee2021crosssum}, (4)~ANT~\cite{chouigui2021arabic}, and (5)~MarSum~\cite{inbook}.

\subsubsection{News Title Generation}

The news title generation (NTG) task is about producing a suitable title for a given news article. That is, a title generation model is required to output a short grammatical sequence of words that are appropriate for the content of the article. For this, we use two datasets: {(1) Arabic NTG}~\cite{nagoudi2022_arat5}, and {(2) XLSum}~\cite{hasan2021xlsum}.\footnote{We note that XLSum~\cite{hasan2021xlsum} has news articles annotated with summaries and titles. We use pairs of articles and titles to create the title generation data.}

\subsubsection{Question Answering}\label{subsec:QA}

For the QA cluster, we use seven publicly available QA datasets across four tasks. %Namely, we use \textit{extractive QA}, \textit{retrieval QA},  \textit{open-domain QA}, and  \textit{multi-choice QA}. 
A summary of the QA cluster is in Appendix Table~\ref{tab:dolphin-stats}. We also provide brief information about each task here. 
% \usepackage{graphicx}
% \usepackage[normalem]{ulem}
\begin{table*}[]
\centering
\resizebox{\textwidth}{!}{%
\begin{tabular}{|l|r|r|}
\toprule
\multicolumn{1}{|l|}{\textbf{Dataset}} & \multicolumn{1}{c|}{\textbf{Source}} & \multicolumn{1}{c|}{\textbf{Target}} \\ \midrule ADT & \<ولما أنهى الكلام على الطهارة المائية صغرى وكبرى انتقل يتكلم على الطهارة الترابية ...> & \<وَلَمَّا أَنْهَى الْكَلَامَ عَلَى الطَّهَارَةِ الْمَائِيَّةِ صُغْرَى وَكُبْرَى انْتَقَلَ يَتَكَلَّمُ عَلَى الطَّهَارَةِ التُّرَابِيَّةِ ...> \\ 
AEC & \<لم أصدق أن صديقي بدأ العمل معي> & \<أين تعملون يا رفاق؟> \\ 
DRG\textsubscript{EGY} & \<باين عليك قلقان. خير حاصل معاك حاجة؟> & \<والله تعبان جدا من ظروف البلد الصعبة> \\ 
DRG\textsubscript{GUL} & \<ما يعجبني أنو أخوي يروح هنا وهناك بدون ما يسوي أي شي مفيد.> & \<يمكن هو يحاول يدور على شغل. هينحل كل شي لمن يلقى واحد.> \\ 
DRG\textsubscript{LEV} & \<ما تنسى تفرشي سنانك قبل ما تنام> & \<ايه هياني قايم> \\ 
TAPACO & \<هي قدمت لي بعض الطعام.> & \<هي ذودتني ببعض الطعام.> \\ 
AraPara & \<لا يسري هذا العرض بالاقتران مع عروض أخرى .> & \<العرض غير صالح مع أي ترقية أخرى.> \\ 
APB & \<سمعت صوت الأسد .> & \<سمعت زئير الأسد .> \\
APGC & \<نوم العوافي ، هل ترغبين من أمك أن تبقى ؟> & \<نوم العوافي ، هل ترغب من أمك أن تبقى ؟> \\ 
% DIA2MSA\textsubscript{EGY} & \<مينفعش تزهق من اللي بتحبه، اللي بيحب حد بجد مش بيزهق منه ده لو بيحب بجد بقي.> & \<لا تستطيع الممل من من تحب الذي بيحب حد بجد لا يمل منة> \\ \midrule
% Arabic NTG & \begin{tabular}[c]{@{}r@{}}\<قال يحيى أبو الفتوح، نائب رئيس مجلس إدارة البنك الأهلى المصرى، اليوم الجمعة،>\\ \<إن مصرفه بدأ تدريجيا، ضخ كميات من البنكنوت الجديد، ليصل إجمالى حجم البنكنوت>\\ \<الجديد الجاهز للضخ حسب احتياجات المواطنين إلى 12 مليار جنيه، لتلبية احتياجات>\\ \<المصريين فى عيد الأضحى المبارك. وأضاف نائب رئيس البنك الأهلى المصرى، فى تصريحات>\\ \<خاصة لليوم السابع، أن البنك وفر النقد الجديد من كل الفئات، مع توزيعه عن طريق شركات توزيع>\\ \<الأموال، وضخه فى شبكة ماكينات الصراف الآلى التى تتجاوز 4500 ماكينة>\\  \<ATM على مستوى الجمهورية، إضافة إلى أكثر من 500 فرع للبنك تنتشر فى كل المحافظات...>\end{tabular} & \<البنك الأهلى: 12 مليار جنيه بنكنوت جديد لتلبية احتياجات المصريين فى العيد> \\ \midrule
MT Darija & \multicolumn{1}{l|}{mzyana had jrida fach tatkon zwina b7al haka} & \multicolumn{1}{l|}{It's nice, this park, when it looks good like that} \\ 
% MT Darija  & \multicolumn{1}{l|}{3lach tkhle3ti?} & \multicolumn{1}{l|}{why did you feel scared?} \\ \midrule
% MT NArabizi & \multicolumn{1}{l|}{\begin{tabular}[c]{@{}l@{}}bonne chance pour le rck vive les koubiéens\\  inchallah tel3a é bonne chance pr alilou\end{tabular}} & \multicolumn{1}{l|}{\begin{tabular}[c]{@{}l@{}}bonne chance pour le RCK vive les koubiéens j'espère \\ la montée et bonne chance pour Alilou\end{tabular}} \\ \midrule
MT EN $\rightarrow$ MSA & \multicolumn{1}{l|}{the report is silent about temporary special measures.} & \<لم يرد في التقرير أي ذكر للتدابير الخاصة المؤقتة.> \\
% MT FR $\rightarrow$ MSA & \multicolumn{1}{l|}{\begin{tabular}[c]{@{}l@{}}Postes financés au moyen des crédits ouverts au titre \\ du personnel temporaire ( autre que pour les réunions ) .\end{tabular}} & \<تمول في إطار المساعدة المؤقتة العامة .> \\ \midrule
CS JO $\rightarrow$ EN & \<يا يغير ال>alloy \<أو يعمله> coating \<منيح> & \multicolumn{1}{l|}{Either he changed the alloy or make it a good coating} \\ \midrule
% CS MA $\rightarrow$ FR & \<اما الشكل بصح بلا انحياز شيرين> charmante \<على اليسا بزاف> surtout\<دابا> & \multicolumn{1}{l|}{\begin{tabular}[c]{@{}l@{}}Mais la forme vraiment sans biais Chirine est beaucoup \\ plus charmante qu'Elissa surtout maintenant\end{tabular}} \\ \bottomrule
\end{tabular}%
}

\caption{Examples from datasets included in \textit{~\dolphin~}. }\label{tab:examples}
\end{table*}

\noindent\textbf{Extractive QA.} We use four publicly available QA datasets: (1) The Arabic QA dataset ARCD~\cite{mozannar2019neural} and the Arabic part of the following three multi-lingual QA test sets: (2) MLQA~\cite{lewis2019mlqa}, (3) XQuAD~\cite{artetxe2020cross}, and (4) TyDiQA~\cite{artetxe2020cross}. For all the extractive QA experiments, we finetune on the  GoldP multilingual  TyDiQA\textsubscript{train}~\cite{artetxe2020cross}  and evaluate on the test sets listed above. 

\noindent\textbf{Retrieval QA.} For this task, we use (5)~LAReQA~\cite{roy-etal-2020-lareqa}, a cross-lingual retrieval QA dataset built by converting the extractive QA dataset XQuAD~\cite{artetxe2020cross} into a retrieval task XQuAD-R. In our benchmark, we focus on the Arabic part of XQuAD-R (AraQuAD-R). 

\noindent\textbf{Open-Domain QA.} In this task, the goal is to answer fact-based questions in natural language. We add (6) DAWQAS, an Arabic \textit{Why} QA dataset~\cite{DAWQAS-2018} to our QA cluster. 

\noindent\textbf{Multi-choice QA.} We also use (7) EXAMS~\cite{hardalov-etal-2020-exams}, a cross-lingual multi-choice QA dataset that covers $26$ languages (including Arabic). 
Since we only have this particular test set for Arabic,  we follow~\newcite{hardalov-etal-2020-exams} in evaluating the models on EXAMS under a zero-shot setting.\footnote{we use the multilingual part for Train and Dev, where no Arabic data is included, and blind-test on the Arabic test split.}

\subsubsection{Question Generation}

The question generation (QG) cluster involves generating a question for a given passage~\cite{gehrmann2021gem}. The model is trained to generate simple questions relevant to passages along with their answers. For this cluster, we use (passage, answer, and question) triplets from five out of the seven QA question datasets described in Section~\ref{subsec:QA}.\footnote{We exclude the multi-choice QA EXAMS~\cite{hardalov-etal-2020-exams}, the open-domain QA DAWQAS~\cite{DAWQAS-2018}. }

\subsubsection{Paraphrase}
The main goal of this task is to produce for a given Arabic sentence a paraphrase with the same meaning. For this, we employ the following four datasets: (1) AraPara, a multi-domain Arabic paraphrase dataset~\cite{nagoudi2022_arat5}, (2)  ASEP, an Arabic SemEval paraphrasing dataset~\cite{cer2017semeval}, (3) Arabic paraphrasing benchmark (APB)~\cite{alian2019towards}, and  (4) the Arabic section of TaPaCo \cite{scherrer-2020-tapaco}, a multilingual paraphrase corpus. 

\subsubsection{Transliteration}
The task of transliteration (TS) is about converting a word or text from one writing system to another while preserving the pronunciation and sound of the original language. We create our TS component using three word-level datasets, as follows: \textbf{(1)~ANETA},  an English-Arabic named entity transliteration and classification dataset proposed by~\newcite{ameur2019anetac}. 
\textbf{(2) ATAR} \cite{Atar-2021}, a word-level parallel corpus containing  human translations between Jordanian Arabizi\footnote{An informal variant of Arabic spoken in Jordan} and MSA. \textbf{(3)~NETransliteration} \cite{merhav2018design}, a bilingual named entity (person names) transliteration dataset mined from Wikidata for English to each of Arabic, Hebrew, Japanese, Katakana, and Russian.  
\subsubsection{Text Rewriting} 

The text rewriting (TR) cluster is about generating a text of the target style while preserving the content of the source input text.  The TR cluster contains two tasks: \noindent{\textbf{(1) DIA $\rightarrow$ MSA.}} This task involves converting a text written in an Arabic dialect into MSA. For this, we use Dial2MSA ~\cite{mubarak2018dial2msa}. Dial2MSA is a parallel dialectal Arabic corpus for converting  Egyptian, Maghrebi, Levantine, and Gulf dialects into MSA.  \noindent{\textbf{(2) Gender Rewriting.}} We use the Arabic parallel gender corpus (APGC) proposed by~\newcite{alhafni-etal-2022-user}, where the task is to take a given input sentence written in one gender (e.g., male) to produce a target sentence that has the same meaning but employing the opposite gender (i.e., female).

\subsubsection{Diacritization}
Arabic text diacritization  (ATD) is the computational process of restoring missing diacritics or vowels to the orthographic word or a sequence of words (i.e., a sentence or a whole text). For this task, we use the Arabic  diacritization dataset proposed by~\newcite{fadel2019arabic}.

\subsubsection{Dialogue Response Generation}\label{subsec:drg}
Dialogue response generation (DRG) is a human-computer interaction task with the goal of automatically producing a human-like response given a dialogue context. In this cluster, we have two tasks:

\noindent{\textbf{(1) MSA DRG.}} For this task, we use the Arabic empathetic chatbot (AEC) dataset~\cite{naous2020empathy}. It contains open-domain utterances with their corresponding empathetic responses machine translated from English into MSA. \noindent{\textbf{(2)  Dialectal DRG.}} We add the open-domain response generation in Arabic dialects proposed by ~\newcite{naous2023open}. Three native translators from the Levantine, Egyptian, and Gulf areas were asked to translate $1$K utterance-response pairs from the English open-domain dialogues dataset DailyDialog~\cite{li2017dailydialog}.

\subsubsection{Grammatical Error Correction}

The task of grammatical error correction (GEC) is focused on analyzing written text, automatically pinpointing, and rectifying a variety of grammatical errors as illustrated by a typical instance of grammatical error correction and its manual rectification. In this cluster, we use three GEC datasets:

\noindent\textbf{(1}-\textbf{2)~QALB.} We use two datasets extracted from the QALB shared tasks from 2014~\cite{mohit2014first} and 2015~\cite{rozovskaya2015second}. Both datasets are manually corrected collections of Arabic texts originating from online commentaries on Aljazeera articles written by native Arabic speakers (L1), as well as texts produced by learners of Arabic as a second language (L2).
\noindent\textbf{(3)~ZAEBUC.} A corpus that focuses on bilingual writers presented by~\newcite{habash-palfreyman-2022-zaebuc}. It matches comparable texts in different languages written by the same writer on different occasions. The corpus is enhanced by adding multiple layered annotations, including manually corrected versions of the raw text, allowing us to use it for GEC.

\subsubsection{Data2Text}
The Data2Text (DT) task involves converting structured data like tables as input into descriptive texts without misrepresenting their contents, while sounding natural in writing (i.e., fluently describing this data as output). For the DT task cluster, we use the Arabic subset of the multilingual dataset MD2T proposed by~\newcite{mille-etal-2020-third} during the third multilingual surface realization shared task.  Table~\ref{tab:examples} shows examples from each task included in ~\dolphin. We now introduce our strong baselines exploiting our benchmark.

% \hl{\textbf{NEXT}}
% \subsection{Text Simplification}

\subsection{Comparative Analysis with ARGEN.}
Compared to the previous largest Arabic NLU benchmark, ARGEN (which we list in Table~\ref{tab:bench_comp}), ~\dolphin~\cite{nagoudi2022_arat5} exhibits several advantages. Specifically, we observe the following:

\noindent\textbf{Coverage.}~\dolphin~boasts a significantly larger dataset pool ($\sim$3X larger). In terms of the number of datasets,~\dolphin~comprises $40$ datasets compared to only $13$ datasets in ARGEN. Hence,~\dolphin~offers a total of $27$ totally new datasets.

\noindent\textbf{Task clusters.}~\dolphin's reach also extends to a wider array of task clusters, encompassing $13$ clusters as opposed to ARGEN's \textit{seven} clusters. \dolphin~introduces \textit{six} novel tasks: \textit{Arabic text diacritization}, \textit{dialogue response generation}, \textit{data-to-text conversion}, \textit{grammatical error correction}, \textit{text rewriting}, and \textit{question answering}.

\noindent\textbf{Availability}.~\dolphin's datasets are drawn exclusively from publicly available sources, while ARGEN involves several non-public datasets such as the machine translation datasets introduced by~\citet{zbib2012machine} and transliteration presented by~\citet{song2014collecting}. As such,~\dolphin~avoids issues ARGEN suffers from such as challenges with (i) public distribution of the data and (ii) ease of evaluation.

\noindent\textbf{Interactivity}.~\dolphin~uniquely offers a benchmark leaderboard, a feature absent in ARGEN, providing real-time performance tracking and a dynamic evaluation environment.

% %%%%%%%%%%%%%%%%%%%%%%

% \usepackage{multirow}
% \usepackage{graphicx}
\begin{table*}[]
\resizebox{\textwidth}{!}{%
\begin{tabular}{lclrrrrr}
\toprule
\textbf{Cluster} & \textbf{Metric} & \multicolumn{1}{l}{\textbf{Test Set}} & \textbf{mT0} & \textbf{mT5} & \textbf{AraBART} & \textbf{AraT5} & \textbf{AraT5v2} \\
\midrule

\multirow{6}{*}{Code-Switching} & \multirow{6}{*}{$Bleu$} &  \textit{Dz-Fr} $\rightarrow$ \textit{Fr}  & 10.90\textsuperscript{$\pm$1.23}	& 11.92\textsuperscript{$\pm$0.91}	& \textbf{18.67}\textsuperscript{$\pm$1.98}	& 12.23\textsuperscript{$\pm$2.32}	& 16.16\textsuperscript{$\pm$1.68}\\ 
 &  &  \textit{Eg-En}  $\rightarrow$ \textit{En}  & \textbf{7.19}\textsuperscript{$\pm$0.45}	& 4.38\textsuperscript{$\pm$1.02} &	1.35\textsuperscript{$\pm$0.65}	& 2.41\textsuperscript{$\pm$0.73}	& 3.22\textsuperscript{$\pm$0.76}  \\ 
 &  & \textit{Jo-En}   $\rightarrow$ \textit{En}    & \textbf{11.37}\textsuperscript{$\pm$1.11} & 8.42\textsuperscript{$\pm$0.87} & 2.0\textsuperscript{$\pm$0.88}	& 4.59\textsuperscript{$\pm$0.32}	& 6.29\textsuperscript{$\pm$0.11}   \\ 
 &  &  \textit{MA-Fr}  $\rightarrow$ \textit{Fr}    &  11.9\textsuperscript{$\pm$0.66}	& 13.63\textsuperscript{$\pm$0.87}	& \textbf{16.14}\textsuperscript{$\pm$0.02}	& 10.87\textsuperscript{$\pm$0.65}	& 14.48\textsuperscript{$\pm$0.32}  \\ 
 &  &  \textit{Ps-En}  $\rightarrow$ \textit{En}    &  \textbf{5.82}\textsuperscript{$\pm$0.87}	& 4.84\textsuperscript{$\pm$0.70}	& 1.170\textsuperscript{$\pm$0.91}	& 2.57\textsuperscript{$\pm$0.51}	& 3.67\textsuperscript{$\pm$0.65} \\ 
 &  & \textit{Ye-En}  $\rightarrow$ \textit{En}    &  \textbf{8.59}\textsuperscript{$\pm$0.07}	& 6.91\textsuperscript{$\pm$0.09}	& 2.8\textsuperscript{$\pm$0.63}	& 3.88\textsuperscript{$\pm$0.76}	& 5.88\textsuperscript{$\pm$0.01}  \\ 
 \midrule
Data2Text & Bleu & MD2T & 0.22\textsuperscript{$\pm$0.02} & 0.17\textsuperscript{$\pm$0.06} & 0.47\textsuperscript{$\pm$0.12} & 0.04\textsuperscript{$\pm$0.01} & \textbf{0.83}\textsuperscript{$\pm$0.22} \\ \midrule
Diacritization & $CER$ & ADT~\textbf{$\downarrow$} & 1.58\textsuperscript{$\pm$0.13}& 1.64\textsuperscript{$\pm$0.11} & 23.43\textsuperscript{$\pm$1.51} & 2.58\textsuperscript{$\pm$0.19} & \textbf{1.36}\textsuperscript{$\pm$0.41} \\ \midrule
\multirow{4}{*}{Dialogue Generation} & \multirow{4}{*}{$Bleu$} & AEC & 1.29\textsuperscript{$\pm$0.21} & 1.14\textsuperscript{$\pm$0.11} & \textbf{1.71\textsuperscript{$\pm$0.03}} & 1.33\textsuperscript{$\pm$0.06} & 1.41\textsuperscript{$\pm$0.24} \\
 &  & DRG\textsubscript{EGY} & 0.05\textsuperscript{$\pm$0.03} & 0.06\textsuperscript{$\pm$0.04} & \textbf{0.35}\textsuperscript{$\pm$0.02} & 0.12\textsuperscript{$\pm$0.03} & 0.32\textsuperscript{$\pm$0.02} \\
 &  & DRG\textsubscript{GUL} & \textbf{1.02}\textsuperscript{$\pm$0.16} & 0.1\textsuperscript{$\pm$0.07} & 0.8\textsuperscript{$\pm$0.33} & 0.29\textsuperscript{$\pm$0.11} & 0.36\textsuperscript{$\pm$0.12} \\
 &  & DRG\textsubscript{LEV} & 0.16\textsuperscript{$\pm$0.11} & 0.11\textsuperscript{$\pm$0.08} & \textbf{0.57}\textsuperscript{$\pm$0.20} & 0.35\textsuperscript{$\pm$0.09} & 0.48\textsuperscript{$\pm$0.13} \\\midrule

% 65.86/0.67	66.45/0.22	68.67/0.08	64.92/0.23	70.54/0.16
% 66.9/0.92	7/0	69.31/1.55	64.22/0.82	70.71/0.61
% 1/0	2/0	3/0	4/0	5/0
% 47.33/3.34	12/0	82.08/7.54	75.78/2.43	84.93/4.46

\multirow{4}{*}{GEC} & \multirow{4}{*}{$F_{0.5}$ ($M$\textsuperscript{$2$})} & QALB 2014 & 65.86\textsuperscript{$\pm$0.67} & 66.45\textsuperscript{$\pm$0.22}	& 68.67\textsuperscript{$\pm$0.08} & 64.92\textsuperscript{$\pm$0.23}& \textbf{70.54}\textsuperscript{$\pm$0.16} \\
 &  & QALB 2015 (L1) & 66.90\textsuperscript{$\pm$0.92}	 & 66.68\textsuperscript{$\pm$0.08} &	69.31\textsuperscript{$\pm$1.55} &	64.22\textsuperscript{$\pm$0.82}	 &\textbf{70.71}\textsuperscript{$\pm$0.61 }\\
 % &  & QALB 2015 L2 & *[] & *[] & *[] & *[] & *[] \\
 &  & ZAEBUC & 47.33\textsuperscript{$\pm$3.34}	& 46.90\textsuperscript{$\pm$0.87}	&82.08\textsuperscript{$\pm$7.54} & 75.78\textsuperscript{$\pm$2.43} &	\textbf{84.93}\textsuperscript{$\pm$4.46}\\ \midrule
\multirow{4}{*}{Paraphrase} & \multirow{4}{*}{$Bleu$} & TAPACO & 15.43\textsuperscript{$\pm$0.64} & 14.89\textsuperscript{$\pm$0.28} & 17.9\textsuperscript{$\pm$1.06} & 15.90\textsuperscript{$\pm$0.06} & \textbf{18.69}\textsuperscript{$\pm$0.26} \\
 % &  & AraPara & 20.04\textsuperscript{$\pm$0.94 & \textbf{22.09\textsuperscript{$\pm$0.85} & 16.21\textsuperscript{$\pm$0.18 & 17.4\textsuperscript{$\pm$0.24 & 15.39\textsuperscript{$\pm$0.33 \\
 &  & APB & \textbf{38.36}\textsuperscript{$\pm$0.14} & 24.29\textsuperscript{$\pm$13.98} & 37.66\textsuperscript{$\pm$1.01} & 20.34\textsuperscript{$\pm$1.82} & 30.18\textsuperscript{$\pm$1.62} \\
 &  & SemEval & 20.49\textsuperscript{$\pm$0.13} & 20.23\textsuperscript{$\pm$0.03} & 24.52\textsuperscript{$\pm$0.62} & 19.33\textsuperscript{$\pm$0.08}& \textbf{27.96}\textsuperscript{$\pm$3.03} \\
 \midrule
\multirow{8}{*}{Question Answering} & \multirow{8}{*}{$F_1$} & LAREQA\textsubscript{QA} & \textbf{63.58}\textsuperscript{$\pm$0.63} & 23.38\textsuperscript{$\pm$1.12} & 45.01\textsuperscript{$\pm$1.98} & 25.45\textsuperscript{$\pm$2.65} & 29.93\textsuperscript{$\pm$4.73} \\
 &  & DAWQS\textsubscript{QA} & 2.52\textsuperscript{$\pm$0.03} & 2.82\textsuperscript{$\pm$0.07} & 4.17\textsuperscript{$\pm$0.30} & 0.37\textsuperscript{$\pm$0.45} & \textbf{4.98}\textsuperscript{$\pm$0.08} \\
 &  & EXAMS\textsubscript{QA} & 42.75\textsuperscript{$\pm$0.61} & \textbf{23.24}\textsuperscript{$\pm$0.55} & 22.54\textsuperscript{$\pm$0.12} & 12.69\textsuperscript{$\pm$0.40} & 28.14\textsuperscript{$\pm$3.80} \\
 &  & MKQA\textsubscript{QA} & 30.01\textsuperscript{$\pm$0.41} & 32.90\textsuperscript{$\pm$0.0} & 32.42\textsuperscript{$\pm$0.09} & 32.9\textsuperscript{$\pm$0.0} & \textbf{33.11}\textsuperscript{$\pm$0.36} \\
 &  & LMQA\textsubscript{QA} & 49.17\textsuperscript{$\pm$0.34} & 45.13\textsuperscript{$\pm$0.35} & 47.24\textsuperscript{$\pm$0.13} & 51.95\textsuperscript{$\pm$0.09 }& \textbf{54.44}\textsuperscript{$\pm$0.56} \\
 &  & ARCD\textsubscript{QA} & 53.24\textsuperscript{$\pm$0.24} & 51.63\textsuperscript{$\pm$1.01} & 50.26\textsuperscript{$\pm$0.99} & 58.12\textsuperscript{$\pm$0.16} & \textbf{61.38}\textsuperscript{$\pm$0.97} \\
 &  & TyDiQA\textsubscript{QA} & 76.31\textsuperscript{$\pm$0.09} &	74.99\textsuperscript{$\pm$0.23} &	73.32\textsuperscript{$\pm$1.21}	& 39.55\textsuperscript{$\pm$1.96}	& \textbf{83.34}\textsuperscript{$\pm$0.45} \\
 &  & XQUAD\textsubscript{QA} & 54.55\textsuperscript{$\pm$0.76}	& 47.43\textsuperscript{$\pm$0.91} &	47.33\textsuperscript{$\pm$0.87}	& 48.71\textsuperscript{$\pm$0.5}	& \textbf{57.88}\textsuperscript{$\pm$0.04}\\\midrule
\multirow{6}{*}{Question Generation} & \multirow{6}{*}{$Bleu$} & LAREQA\textsubscript{QG} & 9.04\textsuperscript{$\pm$0.29} & 5.5\textsuperscript{$\pm$2.99} & \textbf{10.23}\textsuperscript{$\pm$0.72} & 8.65\textsuperscript{$\pm$0.98} & 10.07\textsuperscript{$\pm$0.56} \\
 &  & Arabic-SQUAD\textsubscript{QG} & 9.20\textsuperscript{$\pm$0.07} & 9.01\textsuperscript{$\pm$0.06} & 10.10\textsuperscript{$\pm$0.09} & 8.44\textsuperscript{$\pm$0.11} & \textbf{10.76}\textsuperscript{$\pm$0.18} \\
 &  & MLQA\textsubscript{QG} & 6.04\textsuperscript{$\pm$0.08} & 6.0\textsuperscript{$\pm$0.38}& 7.02\textsuperscript{$\pm$0.09} & 6.12\textsuperscript{$\pm$0.42} & \textbf{7.45}\textsuperscript{$\pm$0.21} \\
 &  & ARCD\textsubscript{QG} & 17.73\textsuperscript{$\pm$0.99} & 17.62\textsuperscript{$\pm$2.10} & \textbf{22.79}\textsuperscript{$\pm$0.66} & 16.8\textsuperscript{$\pm$1.32} & 21.58\textsuperscript{$\pm$1.55} \\
 &  & TyDiQA\textsubscript{QG} & 30.22\textsuperscript{$\pm$0.91} & 31.00\textsuperscript{$\pm$0.97} & \textbf{33.64}\textsuperscript{$\pm$0.13} & 22.09\textsuperscript{$\pm$1.85} & \textbf{33.64}\textsuperscript{$\pm$0.89} \\
 &  & XQUAD\textsubscript{QG} & 10.04\textsuperscript{$\pm$0.01} & 	9.96\textsuperscript{$\pm$0.03} & 	10.27\textsuperscript{$\pm$0.31} & 	9.21\textsuperscript{$\pm$0.09}	 & \textbf{10.82}\textsuperscript{$\pm$0.12} \\ \midrule
\multirow{2}{*}{Text Rewriting} & \multirow{2}{*}{$Bleu$} & APGC & 90.43\textsuperscript{$\pm$0.14} & 90.47\textsuperscript{$\pm$0.04} & 88.93\textsuperscript{$\pm$0.56} & 89.87\textsuperscript{$\pm$0.07} & \textbf{91.19}\textsuperscript{$\pm$0.07} \\
 &  & DIA2MSA\textsubscript{EGY} & 10.35\textsuperscript{$\pm$0.58} & 10.26\textsuperscript{$\pm$0.31} & 12.57\textsuperscript{$\pm$0.27} & 10.53\textsuperscript{$\pm$0.08} & \textbf{14.01}\textsuperscript{$\pm$0.43} \\ \midrule
\multirow{5}{*}{Summarization} & \multirow{5}{*}{$RougeL$} & XLSum & 21.46\textsuperscript{$\pm$0.54} & 20.64\textsuperscript{$\pm$0.31} & 26.64\textsuperscript{$\pm$0.04} & 22.71\textsuperscript{$\pm$1.36} & \textbf{26.88}\textsuperscript{$\pm$0.02} \\
 &  & CrossSum & 21.0\textsuperscript{$\pm$0.38} & 20.29\textsuperscript{$\pm$0.01} & 25.89\textsuperscript{$\pm$0.09} & 22.14\textsuperscript{$\pm$1.53} & \textbf{26.47}\textsuperscript{$\pm$1.02} \\
 &  & MarSum & 23.0\textsuperscript{$\pm$0.17} & 22.57\textsuperscript{$\pm$0.21} & \textbf{26.49}\textsuperscript{$\pm$0.03} & 21.71\textsuperscript{$\pm$0.39} & 25.727\textsuperscript{$\pm$0.02}\\
 &  & MassiveSum & 25.57\textsuperscript{$\pm$0.11} & 22.88\textsuperscript{$\pm$0.12} & \textbf{30.0}\textsuperscript{$\pm$0.11} & 15.89\textsuperscript{$\pm$0.4} &23.07\textsuperscript{$\pm$0.33 }\\
 &  & ANTCorp & 90.29\textsuperscript{$\pm$0.11} & 88.84\textsuperscript{$\pm$0.91} & 90.0\textsuperscript{$\pm$0.2} & 86.64\textsuperscript{$\pm$0.22} & \textbf{91.28 }\textsuperscript{$\pm$0.88} \\ \midrule
\multirow{2}{*}{Title Generation} & \multirow{2}{*}{$Bleu$} & Arabic NTG & 19.03\textsuperscript{$\pm$0.34} & 19.23\textsuperscript{$\pm$0.01} & \textbf{22.75}\textsuperscript{$\pm$0.09} & 19.55\textsuperscript{$\pm$0.16} & 22.27\textsuperscript{$\pm$0.18} \\
 &  & XLSum & 6.50\textsuperscript{$\pm$0.17} & 6.51\textsuperscript{$\pm$0.11} & 8.98\textsuperscript{$\pm$0.18} & 7.44\textsuperscript{$\pm$0.11} & \textbf{9.64}\textsuperscript{$\pm$0.13} \\ \midrule
 
\multirow{3}{*}{Transliteration} & $CER$ & ANTAEC~\textbf{$\downarrow$} & 19.21\textsuperscript{$\pm$0.48} & 18.93\textsuperscript{$\pm$0.30} & 18.29\textsuperscript{$\pm$0.29} & 20.74\textsuperscript{$\pm$0.17} & \textbf{18.44}\textsuperscript{$\pm$0.29} \\
 & $CER$ & ATAR~\textbf{$\downarrow$} & 16.79\textsuperscript{$\pm$0.15} & 16.68\textsuperscript{$\pm$0.22} & 17.70\textsuperscript{$\pm$0.05} & 36.51\textsuperscript{$\pm$1.53} & \textbf{15.20}\textsuperscript{$\pm$0.32} \\
 & $Belu$ & NETTrans & 55.7\textsuperscript{$\pm$0.18} & 55.02\textsuperscript{$\pm$0.47} & 54.15\textsuperscript{$\pm$0.75} & 51.89\textsuperscript{$\pm$0.64} & \textbf{57.41}\textsuperscript{$\pm$0.93} \\ \midrule
\multirow{6}{*}{MT} & \multirow{6}{*}{$Bleu$} & Darija & 16.95\textsuperscript{$\pm$1.81} & 11.27\textsuperscript{$\pm$2.54} & 16.69\textsuperscript{$\pm$0.33} & 1.29\textsuperscript{$\pm$0.46} & \textbf{18.09}\textsuperscript{$\pm$2.85} \\
 &  & NArabizi & \textbf{11.39}\textsuperscript{$\pm$1.84} & 3.37\textsuperscript{$\pm$0.39} & 11.12\textsuperscript{$\pm$1.20} & 6.91\textsuperscript{$\pm$0.01} & 8.98\textsuperscript{$\pm$1.52 }\\
 &  & \textit{En} $\rightarrow$ \textit{MSA} & 23.83\textsuperscript{$\pm$1.04} & 23.68\textsuperscript{$\pm$1.10} & 24.13\textsuperscript{$\pm$0.13} & 22.34\textsuperscript{$\pm$0.13} & \textbf{28.12}\textsuperscript{$\pm$0.24} \\
 &  & \textit{Fr} $\rightarrow$ \textit{MSA} & 17.28\textsuperscript{$\pm$0.71} & 17.74\textsuperscript{$\pm$0.08} & 17.76\textsuperscript{$\pm$0.04} & 15.73\textsuperscript{$\pm$0.12} & \textbf{20.51}\textsuperscript{$\pm$0.10} \\
 &  & \textit{Es}$\rightarrow$ \textit{MSA} & 19.92\textsuperscript{$\pm$0.7} & 20.56\textsuperscript{$\pm$0.06} & 20.38\textsuperscript{$\pm$0.11} & 17.73\textsuperscript{$\pm$0.20} & \textbf{21.74}\textsuperscript{$\pm$0.36} \\
 &  & \textit{Ru} $\rightarrow$ \textit{MSA} & 16.93\textsuperscript{$\pm$0.67} & 17.12\textsuperscript{$\pm$0.18} & 3.46\textsuperscript{$\pm$0.14} & 14.10\textsuperscript{$\pm$0.02} & \textbf{18.29}\textsuperscript{$\pm$0.82} \\ \bottomrule
\textbf{\textit{\dolphin\textsubscript{L}~Score }} &  & \textbf{Avg. $\downarrow$ tasks} & 12.53 & 	12.42 & 	19.81	 & 19.94	 & \textbf{11.67} \\ \midrule
\textbf{\textit{\dolphin\textsubscript{H}~Score}} & & \textbf{Avg. $\uparrow$ tasks} &  	26.32 &	23.88	&26.44 &	22.67	&\textbf{27.82}\\ \bottomrule

\end{tabular}%
}'\caption{Average of three runs of finetuned Arabic and multilingual models on~\dolphin~test. \textbf{\dolphin\textsubscript{L}~Score}: refers to the macro-average scores of tasks where a lower score $\downarrow$ is better. \textbf{\dolphin\textsubscript{H}~Score}:  refers to the macro-average scores of tasks where a higher score $\uparrow$ is better. }\label{tab:test_results}
\end{table*}

\section{Model Evaluation on~\dolphin}\label{sec:Eval}
% \usepackage{multirow}
\begin{table}[t]
\centering
 \renewcommand{\arraystretch}{1.15}
\resizebox{\columnwidth}{!}{%
\begin{tabular}{lcHcccHccc} %%%%%%%%%%%%%%%%%%%%%%%%%%%%%
\toprule
\textbf{Setting}  & \multicolumn{8}{c}{\begin{tabular}[c]{@{}c@{}}\textbf{Few-Shot} \end{tabular}}& \textbf{FFT}\\ 
\cmidrule(r){1-9} \cmidrule(r){10-10} 
\multirow{2}{*}{\textbf{Task}}  & \multicolumn{4}{c}{\begin{tabular}[c]{@{}c@{}}\textbf{BLOOMZ} \end{tabular}}& \multicolumn{4}{c}{\begin{tabular}[c]{@{}c@{}}\textbf{ChatGPT} \end{tabular}}& \multirow{2}{*}{\textbf{AraT5\textsubscript{v2}}} \\ \cmidrule(r){2-5} \cmidrule(r){6-9}
 & $0$  & $3$  & $5$  & $10$ & $0$  & $3$  & $5$  & $10$ & \\ \midrule
CST (Jo-en$\rightarrow$en) & {11.52} & {10.91} & 11.56 & 11.50 & {36.61} & {37.38} &38.55 & \textbf{40.88} & 5.56\\
CST (MSA-fr$\rightarrow$fr)& 28.41 & {28.27} & 26.75 & 28.61 & {34.61} & {35.40} & 36.45 & \textbf{37.95} & 17.49\\

Diacritization~\textbf{$\downarrow$}& 0.51  & 1.33  & 1.62  & 1.42  & 0.11  & 0.06  & 0.05  & 0.06  & \textbf{0.02}\\
Dialogue Generation & 0.00  & 0.28  & 0.38  & 0.44  & 0.38  & 0.49  & 0.51  & 0.00  &  \textbf{0.98}  \\

GEC & 26.42 & 28.78 & 28.13 & -  & 53.59 & 62.41 & 62.04 & -  &  \textbf{96.63}  \\

MT (en$\rightarrow$ar) & 8.33  & 12.54 & 12.35 & 10.07 & 20.52 & 23.58 & 23.34 & 23.74 &  \textbf{26.71}  \\
MT (es$\rightarrow$ar) & 6.94  & 9.20  & 9.31  & 7.33  & 16.47 & 18.11 & 17.45 & 19.32 & \textbf{21.43}\\
MT (fr$\rightarrow$ar) & 6.88  & 5.51  & 5.76  & 4.97  & 15.12 & 15.44 & 15.57 & 16.26 & \textbf{19.11}\\
MT (ru$\rightarrow$ar) & 2.42  & 1.95  & 3.17  & 1.82  & 15.83 & 17.52 & 17.46 & 17.38 &\textbf{18.01} \\

Paraphrase& 12.98 & 9.37  & 10.27 & 10.55 & 7.89  & 8.92  & 9.19  & 9.60  & \textbf{18.90}  \\

Question Answering & 76.04 & 65.45 & 62.08 & 60.49 & 32.98 & 51.73 & 54.14 & 53.67 & \textbf{83.16} \\
Question Generation & 28.76 & 15.70 & 18.53 & 18.69 & 14.48 & 19.86 & 20.08 & 18.15 &\textbf{34.34} \\

Summarization  & 13.56 & 9.13  & 10.74 & 9.63  & 16.88 & 20.01 & 20.40 & 19.58 &\textbf{26.96} \\

Text Rewriting & 76.67 & 23.96 & 13.97 & 12.73 & 41.59 & 58.75 & 53.34 & 62.62 & \textbf{90.75}\\
Title Generation  & 0.99  & 0.79  & 1.20  & 0.62  & 3.24  & 4.72  & 4.62  & 4.54  &\textbf{9.30} \\
Transliteration~\textbf{$\downarrow$} & 0.59  & 0.45  & 0.42  & 0.42  & 0.27  & 0.24  & 0.24  & 0.23  &  \textbf{0.20}  \\

\bottomrule
\end{tabular}}

\caption{
\label{tab:nlg-results}
\textit{K}-shot results with BLOOMZ and ChatGPT, compared to best finetuned model (AraT5\textsubscript{v2}). We report CER for diacritization and transliteration, ROUGE for summarization, F\textsubscript{0.5} (M\textsuperscript{2}) for GEC, and F\textsubscript{1} for QA. All other tasks reported in BLEU. \textbf{$\downarrow$}: lower is better.  }

%Few-shot results with BLOOMZ and ChatGPT, compared to best finetuned model (AraT5\textsubscript{v2}).  \textbf{GEC}: grammatical error correction, \textbf{CST}: code switched translation. We report character error rate (CER) for diacritization and transliteration, ROUGE for Summarization, M2 scorer for GEC and Macro-F\textsubscript{1}  for QA. For all other tasks, we report the BLEU score. Higher is better unless otherwise specified by \textbf{$\downarrow$}.  }

\end{table}

In order to establish a conducive environment for meaningful comparisons on~\dolphin, we offer a number of strong baselines for both finetuning and \textit{k}-shot settings as described next.

\subsection{Finetuned Models} For finetuning, we benchmark five different Arabic and multilingual models on~\dolphin. These are AraT5~\cite{nagoudi2022_arat5}, of which we pretrain a new version that we refer to as AraT5\textsubscript{v2}, AraBART~\cite{eddine2022arabart}, mBART~\cite{liu2020multilingual}, mT5~\cite{xue2020mt5}, and mT0~\cite{muennighoff2022crosslingual}. More information about these models, including our newly introduced AraT5\textsubscript{v2}, is in Appendix~\ref{app:llms}.
% %%%%%%%%%%%%%%%%%%%%%%%%%%%%%%%%%%%%%%%
% \subsection{\hl{Fully Finetuned Evaluations.}}
% %%%%%%%%%%%%%%%%%%%%%%%%%%%%%%%%%%%%

% \moa{In this section, we conduct an evaluation of various Arabic and multilingual S2S LMs on~\dolphin. Specifically, we assess the performance of AraT5~\cite{nagoudi2022_arat5}, AraBART~\cite{eddine2022arabart}, mBART~\cite{liu2020multilingual}, and mT0~\cite{muennighoff2022crosslingual}.\footnote{We provide a description of each of these models in the Appendix~\ref{sec:llms}.}}

\noindent For all models, we finetune on the training data split (Train) for $20$ epochs with an early stopping of $5$ epochs, learning-rate of  $5e-5$, batch size of $16$, and sequence length of $512$.\footnote{Except for GEC, where we use a seq length of $1,024$.} For all the experiments, we  identify the best model on the respective development split (Dev) and blind testing on the test split (Test). We methodically evaluate each task cluster, ultimately reporting a single \textit{\dolphin~score} following e.g.,~\newcite{wang2018glue} and~\newcite{elmadany2022orca}. \textit{Dolphin~score}  is simply the macro-average of the different scores across task clusters, where each task is weighted equally. Since some of our tasks are reported in metrics where lower numbers are better, we split our metric into \textbf{\textit{\dolphin\textsubscript{L}~score}} (for tasks where lower $\downarrow$  is better [i.e., CER]), and \textbf{\textit{\dolphin\textsubscript{H}~score}} (for tasks where higher  $\uparrow$ is better [i.e., BLEU, F\textsubscript{1}, M\textsubscript{2}, and ROUGE]). Table~\ref{tab:test_results} presents the results of all pretrained models on each task cluster of Dolphin independently using the relevant metric.

\noindent\textbf{Discussion.} As Table~\ref{tab:test_results} shows, models dedicated to Arabic outperform multilingual models on tasks where higher is better (in~\textit{\dolphin\textsubscript{H}}). We also note that AraT5\textsubscript{v2} the new model we build on top of~\cite{nagoudi2022_arat5}, achieves the best \dolphin\textsubscript{H} and \dolphin\textsubscript{L}, at $27.82$ and $11.67$, respectively. It is followed by AraBART with  \dolphin\textsubscript{H} of $26.44$, where a higher score indicates better performance.  Conversely, mT5 achieves a \dolphin\textsubscript{L} of $12.42$, which is considered better in the opposite scenario. We also note that  \textbf{AraT5\textsubscript{v2}} achieves the best results in $30$ individual tasks out of $50$, followed by  AraBART and mT0, where each one excels in $11$ and $8$ individual tasks, respectively.\footnote{We investigate why AraT5 achieves worst, in spite of being dedicated to Arabic, finding it to perform better with $100$ epochs and a patience of $20$ as~\newcite{nagoudi2022_arat5} report.}

\begin{figure}[t]
 \includegraphics[scale=0.32,left]{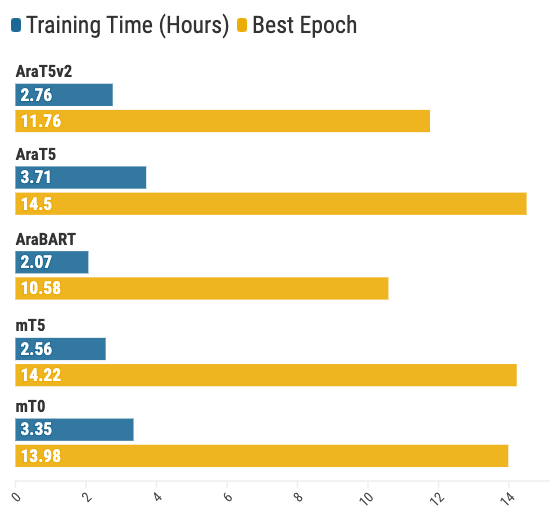}
  \caption{Finetuning time (in hrs) and no. of epoch. We report the average of three runs across all tasks. } \label{fig:cost} 
 \end{figure}

\paragraph{Model Computational Costs.}
We assess the computational efficiency of the Arabic and multilingual models we finetune.  Figure~\ref{fig:cost} shows for each model the \textit{total time needed for convergence} (under our $20$ epochs constraint with a patience of $5$) and the conversion epoch. AraBART is  the fastest ($2.07$ hours), with an average of $10.58$ epochs to convergence, followed by mT5, AraT5\textsubscript{v2}, mT0, and finally AraT5.

% \noindent\textbf{Discussion.} 

% \hl{We need to discuss the results here. }

\subsection{Few-Shot Evaluation.} We also carry out \textit{k-}shot evaluations of both BLOOMZ\footnote{BLOOMZ is finetuned on multiple tasks in $46$ languages, including $\sim1\%$ Arabic.} (7.1B)~\cite{muennighoff2022crosslingual} and ChatGPT (gpt-3.5-turbo)\footnote{
We evaluate the version existing on March 1st, 2023.} on $12$ different NLG tasks across $16$ test sets extracted from~\dolphin.\footnote{We only exclude the data-to-text task.} To keep the cost manageable, we randomly sample a set of $200$ examples from the test set of each task for evaluation. We then 
 evaluate both models under \textit{0}-, \textit{5}-, and \textit{10}-shot settings.  For all experiments, we set the temperature to \textit{zero} to generate deterministic and reproducible results.  We compare both models' performance to our best fully finetuned model, AraT5\textsubscript{v2}, blind-tested on the same sampled $200$ examples. 
 
 \noindent\textbf{Discussion.} Tables~\ref{tab:nlg-results}, shows that ChatGPT outperforms  BLOOMZ in all the $16$ NLG tasks under \textit{0}-, \textit{5}-, and \textit{10}-shot settings. The only exception is the text rewriting task in the 0-shot setting. It is worth mentioning that AraT5\textsubscript{v2} outperforms both ChatGPT and BLOOMZ by $14$ out of $16$. However, ChatGPT (\textit{10}-shot) achieves the highest score in both code-switching tasks, perhaps due to its multilingual pretraining data.

%%%%%%%%%%%%%%%%%%%%%%%%%%%
%%%%%%%%%%%%%%%%%%%%%%%%%%%
%In Figure xx (Appendix xx), we present the total time (in minutes) required for convergence for each model, while Figure E.1 (Appendix E) displays the average number of epochs (out of 20 epochs) until convergence. According to Figure XX, yyyy is the fastest model, \hl{bla bla bla bla}. }
%%%%%%%%%%%%%%%%%%%%%%%%%%%%%%%%%%%

% %%%%%%%%%%%%%%%%%%%%%%
% \input{disc}
% %%%%%%%%%%%%%%%%%%%%%%
\section{Conclusion}\label{sec:conc}

We presented \dolphin, a large and diverse benchmark for Arabic NLG composed of $40$ datasets that are arranged in $13$ tasks.~\dolphin~is designed to facilitate meaningful comparisons and encourage healthy competition in Arabic. We also provide an interactive leaderboard with a range of useful tools and detailed metadata to help situate future research in a rich context of information sharing.~\dolphin~datasets are all publicly available, which should facilitate the adoption and further development of the benchmark. In the future, we intend to build on top of~\dolphin~by extending it to more tasks and Arabic varieties.% Moreover, we have the capability to expand and modify our benchmark in the future to encompass additional Arabic and multilingual datasets. This will aid in their acceptance and utilization by a larger research community.

% %%%%%%%%%%%%%%%%%%%%%%
%\input{Acknow}

% %%%%%%%%%%%%%%%%%%%%%%
% \newpage 
\section{Limitations}\label{sec:limitations}
In spite of the diversity, wide-coverage, high-quality datasets, accessibility, and challenging nature of~\dolphin, it is not without limitations. In particular, we identify the following limitations.

\begin{enumerate}
    \item \textbf{Coverage of Arabic Varieties.} While we make efforts to incorporate tasks from all Arabic varieties, it is important to note that there is a lack of available downstream datasets from countries such as Djibouti, Mauritania, and Yemen. Consequently, these varieties are not currently included in~\dolphin. We hope that the community will develop resources representing all Arab countries, including these, across the various tasks. We also hope that future versions of our benchmark will have extended dialectal coverage in ways that enhance its representation of the Arabic language and help foster technological inclusion.

\item \textbf{Machine-Translated Datasets.}~\dolphin~includes two machine-translated data, AEC~\cite{naous-etal-2021-empathetic} and AraPara~\cite{nagoudi2022_arat5}). While these datasets increase task coverage in~\dolphin, the MT process may inadvertently introduce some biases. For example, MT can result in a narrow representation of language patterns and structures, leading to a limited understanding of the complexities and nuances of different languages. Additionally, benchmark datasets may not adequately capture the wide range of domains, genres, and styles that exist in real-world translation scenarios. This can limit the generalizability of models trained on such data, as they may struggle to handle unfamiliar or specialized content. We hope that future versions of~\dolphin~will involve real-world data that further complement (or even substitute) these translated datasets.

\item \textbf{Automated Evaluation.} Although all NLP depends heavily on automated evaluation to speed up model development, automated methods have their limitations, especially for some tasks. That is, in addition to automated evaluation, some tasks may need human evaluation. In particular, we believe human evaluation can play a crucial role in NLG tasks such as open-domain dialogue generation. For example, it can capture the nuanced aspects of dialogue quality, such as coherence, relevance, and appropriateness. In addition, human evaluation can allow for a comprehensive assessment of the generated dialogues, taking into account contextual understanding, fluency, and overall user experience. This feedback is invaluable in refining and improving dialogue generation models, ensuring that they meet the high standards of human-like conversation.
\end{enumerate}
\section{Ethics Statement}\label{sec:ethics}
\textbf{Data Collection and Release.} \dolphin~is based on publicly available datasets that would not be possible without the hard work of a large number of researchers over the years. We are grateful for these efforts invested by pioneer colleagues. One downside of benchmarking could be that the original authors of the different datasets are not sufficiently acknowledged. In our work, we make sure that all publications of resources we use are properly cited, both by referencing these in this paper (Section~\ref{sec:dolphin}) and highlighting them in our GitHub and leaderboard website. 
\begin{enumerate}
    \item \noindent\textbf{Data Privacy.} Regarding data involved in~\dolphin, we develop the benchmark using publicly available  data. For this reason,  we do not have significant privacy concerns. In addition, the new datasets we develop and release for code-switched machine translation have undergone manual inspection to ensure there is no unintended leak of privacy information in any of the samples.

    \item \noindent \textbf{Intended Use.} We believe our work will spur further research on studying language models on Arabic NLG benchmark. We create a publicly available leaderboard and benchmark several multilingual and Arabic-dedicated SOTA models on \dolphin. The benchmark will facilitate a unified evaluation and pave the way for a healthy competition that could push SoTA on Arabic language generation. 

    \item \noindent \textbf{Potential Misuse and Bias.}
The datasets we collect to create Dolphin may contain potential harmful contents. Additionally, the models we evaluate might be exposed to bias and as a result may generate unintended contents. Therefore, we recommend that these datasets and models not be used in applications without careful prior consideration of potential misuse and bias.
\end{enumerate}

% %%%%%%%%%%%%%%%%%%%%%%
\section*{Acknowledgements}\label{sec:acknow}
We gratefully acknowledge support from Canada Research Chairs (CRC), the Natural Sciences and Engineering Research Council of Canada (NSERC; RGPIN-2018-04267), the Social Sciences and Humanities Research Council of Canada (SSHRC; 435-2018-0576; 895-2020-1004; 895-2021-1008), Canadian Foundation for Innovation (CFI; 37771), Digital Research Alliance of Canada,\footnote{\href{https://alliancecan.ca}{https://alliancecan.ca}} UBC ARC-Sockeye.\footnote{\href{https://arc.ubc.ca/ubc-arc-sockeye}{https://arc.ubc.ca/ubc-arc-sockeye}} We thank the
Google TFRC program for providing us with free TPU access.\footnote{\href{https://sites.research.google/trc/about/}{https://sites.research.google/trc/about/}}

%Any opinions, conclusions or recommendations expressed in this material are those of the author(s) and do not necessarily reflect the views of CRC, NSERC, SSHRC, CFI, CC, AMD, Google, or UBC ARC-Sockeye.

\normalem
% \clearpage
% \newpage
\bibliography{anthology,custom}
\bibliographystyle{acl_natbib}

\appendix

%%%%%%%%%%%%%%%%%%%%%%%
% #############################################################
\clearpage
\appendixpage
\addappheadtotoc
\numberwithin{figure}{section}
\numberwithin{table}{section}
%%%%%%%%%%%%%%%%%%%%%%%%%%%%%%%%%%%%%%%%%%%%%%%%%%%
% \section*{Appendix}
% \label{app:appendix}

We organize our appendices as follows:\\

\noindent \textbf{Sections list}:
\begin{itemize}\setlength\itemsep{.2em}
    \item NLG Benchmarks. (Section~\ref{app:RW_app})
    \begin{itemize}
        \item Arabic (Subsection~\ref{app:RL_ar})
        \item X-Specific (Subsection~\ref{app:RL_x})
        \item Multi-Lingual (Subsection~\ref{app:RL_ml})
     \end{itemize}
     \item Dolphin Tasks (Section~\ref{app:dolphin_tasks})
     \item S2S LLMs (Section~\ref{app:llms})
     % \item \textcolor{red}{Examples from Each Task. (Section~\ref{app:examples})}
     \item Public Leaderboard. (Section~\ref{app:leaderboard}
     
\end{itemize}

\begin{figure*}[!ht]
 \includegraphics[scale=0.18,left]{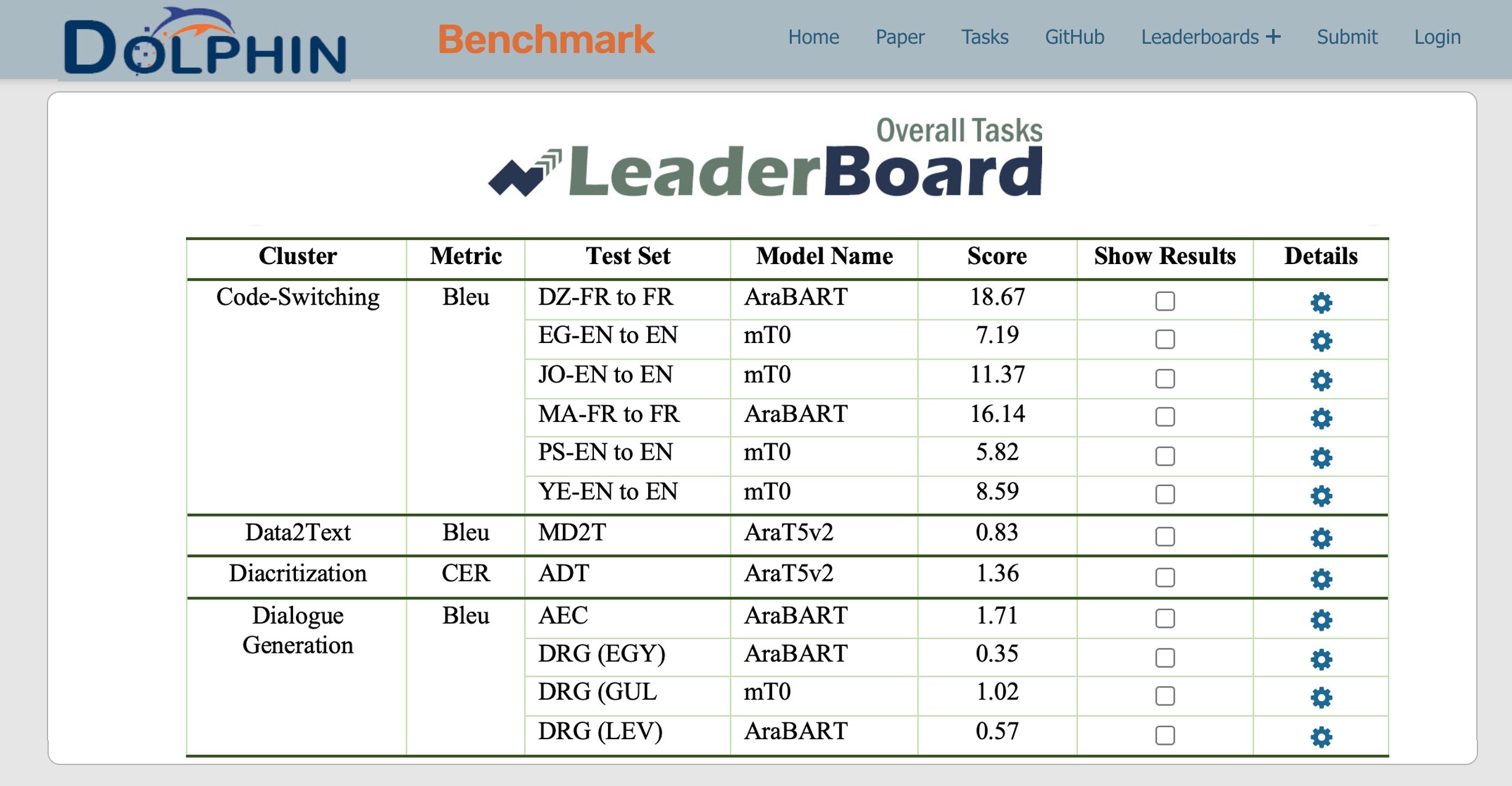}
  \caption{\dolphin~main leaderboard allows showing detailed scores by all models for a given task. } \label{fig:leaderboard} 
 \end{figure*}

\noindent \textbf{Tables and Figures List}:

\begin{itemize}
    \item Statistics of our Dolphin benchmark across the different task clusters (Table~\ref{tab:dolphin-stats}).
    \item \dolphin's Leaderboard (Figure~\ref{fig:leaderboard})
    % \item \textcolor{red}{Examples from Each Task. (Table~\ref{tab:examples})}
\end{itemize}

% \section*{Appendices}
% \onecolumn
% \appendixpage
% \addappheadtotoc

\section{NLG Benchmarks}\label{app:RW_app}
    Existing NLG benchmarks can be classified into three distinct categories: \textit{Arabic-specific}, \textit{X-specific} (where X refers to languages other than Arabic, such as English, Chinese, and others), and \textit{multilingual} benchmarks. In this section, we shall provide a brief overview of each category, highlighting their respective characteristics and scope. We will highlight aspects such as the target language, dataset size, and the breadth of tasks covered. This analysis is summarized in Table~\ref{tab:bench_comp} and Figure~\ref{fig:bench_comp}. The current NLG benchmarks can be divided into three main groups: benchmarks that focus on Arabic, benchmarks that focus on languages other than Arabic (X-specific), and benchmarks that cover multiple languages. In this section, we will give a brief summary of each category, emphasizing their unique features and scope. We will discuss factors like the target language, dataset size, and the range of tasks included.
 
% \subsection{ Benchmarks}

% \begin{itemize}
%     \item \noindent\textbf{GLGE}\cite{liu2020glge} The General Language Generation Evaluation (GLGE) benchmark is a multi-task benchmark for assessing NLG in the English language generalization abilities. It has eight English language generation tasks, including dialogue, generative question answering, question generation, and text summarization.
% \end{itemize}
% \textit{data-to-text}, \textit{dialog response generation}, \textit{generative question answering}, \textit{question generation},   \textit{simplification}, and \textit{summarization}.

% propose SuperGLUE, a benchmark styled after GLUE with a new set of more challenging tasks. SuperGLUE is built around eight tasks and  arranged into four task clusters: QA, NLI, WSD, and coreference resolution. The benchmark is accompanied by a leaderboard with a single-number performance metric (i.e., the \textit{SuperGLUE score}). 

\subsection{Arabic Benchmarks} \label{app:RL_ar}
 \noindent \textbf{AraBench.}  AraBench is an evaluation benchmark  for dialectal Arabic to English machine translation (MT) introduced by~\cite{sajjad2020arabench}. It consists of five publicly available datasets: Arabic-Dialect/English Parallel Text (APT)~\cite{zbib2012machine}, Multi-dialectal Parallel Corpus of Arabic (MDC)~\cite{bouamor2014multidialectal}, MADAR Corpus~\cite{bouamor2018madar}, Qatari-English speech corpus~\cite{elmahdy2014development}, and the English Bible translated into MSA, Tunisian, and Morocco.\footnote{The United Bible Societies https://www.bible.com}

 \noindent\textbf{\textit{AraOPUS-20.}} This is an MT benchmark proposed by~\newcite{nagoudi-2022-turjuman}. It consists of parallel bitext between Arabic and $20$ languages extracted from the OPUS publicly available corpora~\cite{OPUS}. The languages paired with Arabic include high-resource languages such as \textit{English, French}, and \textit{Spanish} and low-resource ones such as \textit{Cebuano},\footnote{Language spoken in the southern Philippines} \textit{Tamashek},\footnote{\textit{Tamashek} is a variety of Tuareg, a Berber macro-language widely spoken by nomadic tribes across North Africa countries.} and \textit{Yoruba}.\footnote{Yoruba is a language spoken in West Africa, primarily in Southwestern Nigeria.}

\noindent\textbf{ARGEN.} The \textbf{AR}abic natural language \textbf{GEN}eration~(\textbf{ARGEN}) benchmark was introduced by~\newcite{nagoudi2022_arat5}. It is composed of $19$ datasets and covers the seven tasks: machine translation, code-switched text translation, summarization, news title generation, question generation, paraphrasing, and transliteration.  

\subsection{X-Specific Benchmarks}\label{app:RL_x}

\noindent\textbf{GLGE.} The \textbf{G}eneral \textbf{L}anguage \textbf{G}eneration \textbf{E}valuation(GLGE) by~\newcite{liu2020glge} is a multi-task benchmark for evaluating the generalization capabilities of NLG in the English language. GLGE has eight English language generation datasets, covering four  NLG tasks:  data-to-text, dialog, table-to-text, and summarization.

%\hl{https://arxiv.org/abs/2104.08200}
%IndoNLG, BanglaNLG, CUGE, CLSE.

\noindent\textbf{BanglaNLG.} BanglaNLG is a benchmark designed for Bangala~\newcite{bhattacharjee2022banglanlg} comprising seven datasets across six NLG tasks:  machine translation, text summarization, question answering, dialogue generation, headline generation, and cross-lingual summarization.

%\hl{https://arxiv.org/abs/2205.11081}

\noindent\textbf{CUGE.} The \textbf{C}hinese Language \textbf{U}nderstanding Generation \textbf{E}valuation \textbf{B}enchmark \newcite{yao2021cuge} covers both language understanding and generation. The language generation collection contains nine datasets across eight tasks. The tasks are  open-domain question answering, document retrieval, summarization, data-to-text, knowledge-driven conversation,  machine translation,  cross-lingual text summarization, and  mathematical computation. The benchmark also covers the tasks of grammatical error correction and reverse dictionary generation, but treats these under the NLU component.  

\noindent\textbf{Bahasa Indonesia.} The Bahasa Indonesia language has over 200M active speakers, yet it is still considered a low-resource language. To overcome this problem,~\cite{guntara-etal-2020-benchmarking} introduced a machine translation benchmark with 14 datasets across four domains: news, religion, conversation, and general.  

\noindent\textbf{PhoMT.} \newcite{doan-etal-2021-phomt} introduces a new Vietnamese-English parallel dataset that is larger and of higher quality than the existing benchmark corpus. The authors conduct experiments to evaluate various translation models on the new dataset and find that the best performance is achieved by fine-tuning the pre-trained sequence-to-sequence denoising auto-encoder mBART.

\noindent\textbf{LOT.}  The \textbf{LO}ng \textbf{T}ext understanding and generation benchmark targets Chinese long text modeling in a story-centric manner~\newcite{10.1162/tacl_a_00469}. LOT combines two comprehension tasks and two-generation tasks. The two generation tasks are commonsense reasoning and discourse structure.
%\hl{https://arxiv.org/abs/2112.13610}

%%%%%%%%%%%%%%%%%%%%%%%%%%%%%%
%%%%%%%%%%%%%%%%%%%%%%%%%%%
 % \begin{figure}[t]
 % \includegraphics[scale=0.32,left]{Image/Bench_Compar7.png}
 %  \caption{A compassion of the number of datasets and tasks supported by the    Arabic (including \textit{\dolphin}), X-specific, and Multilingual NLG benchmarks. } \label{fig:bench_comp} 
 % \end{figure}
%%%%%%%%%%%%%%%%%%%%%%%%%%%
%%%%%%%%%%%%%%%%%%%%%%%%%%%%%%%
\subsection{Multi-Lingual NLG Benchmarks}\label{app:RL_ml}

\noindent\textbf{IndoNLG.} IndoNLG covers three low resources languages widely spoken in Indonesia: Indonesian, Javanese, and Sundanese \newcite{cahyawijaya2021indonlg}. It consists of ten distinct datasets, encompassing four tasks. These are  summarization, question answering, chit-chat, %and three machine translation pairs.
and machine translation.

\noindent\textbf{CLSE.}
The \textbf{C}orpus of \textbf{L}inguistically \textbf{S}ignificant \textbf{E}ntities \newcite{chuklin2022clse} is a multilingual named entities corpus that covers $34$ languages, $74$ semantic classes, and $222$ distinguishable linguistic signatures. The authors also developed an expanded version of the Schema-Guided Dialog Dataset (SG-CLSE) to illustrate one of the potential uses of CLSE in three languages: French, Marathi, and Russian.

%\hl{https://arxiv.org/abs/2211.02423}

\noindent\textbf{GEM\textsubscript{v1}.} The \textbf{G}eneration \textbf{E}valuation and \textbf{M}etrics benchmark~\cite{gehrmann2021gem}
 is a multi-lingual benchmark environment for NLG.  GEM features $18$ languages across $13$ datasets  spanning five NLG tasks: data-to-text, dialog response generation, reasoning, summarization, and simplification.\footnote{Two of the datasets do not include English at all.}

\noindent\textbf{GEM\textsubscript{v2}.}~\newcite{gehrmann2022gemv2} propose a second version, GEM\textsubscript{v2}, styled after GEM\textsubscript{v1} with a new set of datasets and more challenging tasks. This new version supports $40$ documented datasets in $51$ languages. It introduces a modular infrastructure for datasets and models, with an online evaluation process that collects model outputs and computes metrics for all datasets. GEM\textsubscript{v2} is built around nine NLG tasks data-to-text, dialog response generation, paraphrasing, generative question answering, question generation, reasoning, slide generation, simplification, and summarization. %More information about GEM\textsubscript{v1} and GEM\textsubscript{v2} datasets can be found in Appendix~\ref{app:gem}.

\noindent\textbf{IndicNLG.} The first benchmark for Indic languages~\newcite{kumar-etal-2022-indicnlg} covers $11$ Indic languages belonging to two language families: Indo-Aryan and Dravidian. IndicNLG involves the five following tasks: biography generation, news headline generation, sentence summarization, paraphrase generation, and question generation.

\noindent\textbf{MTG.}~\newcite{chen-etal-2022-mtg} introduce  the \textbf{M}ultilingual \textbf{T}ext \textbf{G}eneration to promote knowledge transfer and cross-lingual generation between arbitrary language pairs. MTG contains $400$K of humanly annotated  data samples in five languages, covering four generation tasks. These are story generation, question
generation, title generation, and text summarization.

%\noindent\textbf{GLUECoS.}

%\hl{https://aclanthology.org/2022.emnlp-main.360/}
% \input{Tables/NLG_Bench_Comp}

% \input{Tables/dolphin_tasks}

\section{Dolphin Tasks} \label{app:dolphin_tasks}

\begin{table*}[]
\centering
 \renewcommand{\arraystretch}{1}
% \resizebox{\columnwidth}{!}{%
\resizebox{0.95\textwidth}{!}{%
\caption{Example Table}
\begin{tabular}{|l|l|l|l|c|c|c|c|}
\toprule
\textbf{Task Cluster} & \textbf{Task}  & \textbf{Test Set} & \textbf{~~~~~~~~~~~~~Source} &  \textbf{Train$^\star$ }& \textbf{Dev$^\dagger$} & \textbf{Test$^\ddagger$} \\
\toprule
\multirow{10}{*}{\textbf{MT}}  & \multirow{4}{*}{\textit{X $\rightarrow$ MSA}}  & \textit{De $\rightarrow$ Ar } & \multirow{1}{*}{\newcite{eisele2010multiun}{$^\star$}}  &  \textit{ $50$K}  & $4$K & $4$K \\
                    &  & \textit{En $\rightarrow$ Ar } &  \multirow{3}{*}{\newcite{ziemski2016united}$^\dagger$$^\ddagger$} & \textit{ $50$K}  & $4$K & $4$K \\
                    &  & \textit{Fr $\rightarrow$ Ar } & & \textit{ $50$K}   & $4$K & $4$K\\
                    &  & \textit{Ru $\rightarrow$ Ar }  &  &  \textit{ $50$K} & $4$K & $4$K \\\cline{3-7}
                    % &  & \textit{Zh $\rightarrow$ Ar } & &   & $4$K & $4$K \\ 
                   &  \multirow{2}{*}{\textit{Arabizi $\rightarrow$ X}} &\textit{Dz $\rightarrow$ Fr } & \newcite{seddah-etal-2020-building}  & $1.1$K & $144$ & $146$ \\ 
                    &   & \textit{Ma $\rightarrow$ En } &\newcite{outchakoucht2021moroccan}  & $8$K & $2$K & $2$K \\\cline{3-7}
                    & \multirow{5}{*}{\textit{DA $\rightarrow$ En  }}   
                    & \textit{Eg $\rightarrow$ En } & \multirow{4}{*}{{\newcite{nagoudi-2022-turjuman}$^\star$}}
                     & \multirow{5}{*}{ \textit{$50$K}}  & $200$ & $800$  \\
                    &  & \textit{Jo $\rightarrow$ En }   &  & &$200$ & $800$  \\
                    & & \textit{Ps $\rightarrow$ En }   & \multirow{2}{*}{{\newcite{bouamor2014multidialectal}$^\dagger$$^\ddagger$}} & &$200$ & $800$  \\
                    & &    \textit{Sy $\rightarrow$ En }&   & &$200$ &$800$  \\
                    & & \textit{Tn $\rightarrow$ En } &  &  &$200$ & $800$  \\\midrule

\multirow{6}{*}{\textbf{Code-Switching}}  &  \multirow{6}{*}{\textit{DA-X~$\rightarrow$~X}} & \textit{ Dz-Fr~$\rightarrow$~Fr} & \multirow{5}{*}{{\newcite{nagoudi-2022-turjuman}$^\star$}}  &  \multirow{6}{*}{$50$K}  & $50$ & $250$  \\ 
                 &   & \textit{Ma-Fr~$\rightarrow$~Fr}&   & & $50$ & $250$ \\ 
                   &   & \textit{Tn-Fr~$\rightarrow$~Fr}  &\multirow{3}{*}{\textcolor{blue}{\textit{Our work}}$^\dagger$$^\ddagger$}  &  & $50$ & $250$  \\% \cline{3-7}

                    &  & \textit{Eg-En~$\rightarrow$~En }  & & &  $50$ & $250$  \\ 
                    &   & \textit{Jo-En~$\rightarrow$~En} &  &  & $50$ & $250$  \\ 
                    &   & \textit{Ps-Fr~$\rightarrow$~En} &  &  & $50$ & $250$  \\ 
                    \midrule %\multirow{3}{*}{\textit{DA-En~$\rightarrow$~En}} ### \multirow{3}{*}{$100$K} 
\multirow{5}{*}{\textbf{Summarization}}  &  \multirow{4}{*}{\textit{MSA $\rightarrow$ MSA}} 
            & \textit{ANT Corpus}  & \newcite{chouigui2021arabic} &  $25.2$K     & $3.1$K & $3.1$K  \\ 
            &   & \textit{CrossSum}  & \newcite{bhattacharjee2021crosssum} &  $37.3$K     & $4.6$K & $4.7$K  \\ 
            &  &   \textit{MassiveSum}  & \newcite{varab-schluter-2021-massivesumm} &   $4.6$K     & $459$ & $1.3$K  \\ 
                    &   & \textit{XLSum}  & \newcite{hasan2021xlsum} &  $37.5$K     & $4.7$K & $4.7$K  \\

                    \cline{3-7}
                    & \textit{DA $\rightarrow$ DA}&  \textit{MarSum}    & \newcite{inbook} & $16$K  & $1.7$K  & $1.9$K  \\ \midrule

\multirow{2}{*}{\textbf{Title Generation}}  & \multirow{2}{*}{ \textit{MSA $\rightarrow$ MSA }}   &     \textit{Arabic NTG}  & \newcite{nagoudi2022_arat5} & $93.3.5$K     & $11.6$K & $11.6$K  \\ 
                    &  &    \textit{XLSum}   &  \newcite{hasan2021xlsum} & $37.5$K  & $4.7$K  & $4.7$K  \\ \midrule

\multirow{7}{*}{\textbf{QA/QG}}  & 
                    \multirow{7}{*}{\textit{ MSA $\rightarrow$ MSA  }}  & \textit{ARCD} &\newcite{mozannar2019neural}$^\dagger$$^\ddagger$&   $49.9$K  & $693$ & $702$ \\
                    &  & \textit{MLQA}&  \newcite{lewis2019mlqa}$^\dagger$$^\ddagger$  & $49.9$K & $517$ & $5.3$K \\
                    &  & \textit{XQuAD}& \newcite{artetxe2020cross}$^\ddagger$   & $49.9$K & $5.08$K & $1.1$K\\
                    &  & \textit{TyDiQA} & \newcite{artetxe2020cross}$^\star$$^\ddagger$ & $49.9$K   & $5.08$K & $921$ \\ 

                    &   & \textit{LAReQA}  &\newcite{roy-etal-2020-lareqa} &  $851$ & $119$ & $220$ \\
                    &  & \textit{DAWQAS}  & \newcite{DAWQAS-2018} & $2.2$K & $318$ & $645$ \\
                    &   & \textit{EXAMS}  & \newcite{hardalov-etal-2020-exams} & $7.9$K & $2.6$K & $13.5$K \\ \midrule

 %\textit{Retrival }  # \textit{Open-Domain } # \textit{Multi-choice }
\multirow{3}{*}{\textbf{Transliteration}}  &  
            \multirow{3}{*}{\textit{Arabizi  $\rightarrow$ MSA}} &  \textit{ANETAC}  & \newcite{ameur2019anetac} &   $75.9$K     & $1$K & $3$K  \\ 
                    &   & \textit{ATAR}  & \newcite{Atar-2021} &  $17.2$K     & $2.1$K & $2.1$K  \\ 
                    &  & \textit{NETTrans.}   & \newcite{merhav2018design} & $116$K  & $14.5$K  & $14.5$K  \\ \midrule

\multirow{5}{*}{\textbf{Text Rewriting}}  & \multirow{4}{*}{{\textit{DA $\rightarrow$ MSA}}}   &   \textit{Egy $\rightarrow$ MSA } &  \multirow{4}{*}{{{\newcite{mubarak2018dial2msa}}}}  &  $3.8$K     & $551$ & $1.1$K  \\

                    &   & \textit{Mag $\rightarrow$ MSA } & &   $3.4$K     & $491$ & $996$ \\
                    &   & \textit{Lev $\rightarrow$ MSA } &  &   $4.2$K     & $594$ & $1.2$K \\
                    &  & \textit{Gul $\rightarrow$ MSA }  &  &  $4.2$K     & $594$ & $1.2$K \\ \cline{3-7}

                    & MSA $\rightarrow$ MSA  &  \textit{APGC}  &  \newcite{alhafni-etal-2022-user} & $40.4$K  & $4.7$K  & $11.3$K

                    \\ \midrule

\multirow{1}{*}{\textbf{Diacritization}}  & CA $\rightarrow$ CA &     \textit{ATD}  & \newcite{fadel2019arabic} &  $50$K     & $2.5$K & $2.5$K  \\\midrule
\multirow{1}{*}{\textbf{Data2Text}}  & Table $\rightarrow$ MSA  &     \textit{MD2T}  & ~\newcite{mille-etal-2020-third} &  $6$K     & $900$  & $680$  \\\midrule

\multirow{3}{*}{\textbf{Dialogue Generation}}  & 

MSA$\rightarrow$ MSA  &     \textit{AEC}  & \newcite{naous2020empathy}  & $32.9$K  & $1.8$K  & $1.8$K  \\  \cline{3-7}

& \multirow{3}{*}{\textit{DA $\rightarrow$ DA}}  &   \textit{Egy $\rightarrow$ Egy} & \multirow{3}{*}{\newcite{naous2023open}}  &  $2.1$K     & $297$ & $600$  \\ 
                                          &    & \textit{Lev $\rightarrow$ Lev}   & &  $2.1$K     & $297$ & $600$  \\ 
                                            &    &   \textit{Gul $\rightarrow$ Gul}  & &  $2.1$K     & $297$ & $600$  \\ 
                       \midrule

\multirow{3}{*}{\textbf{GEC}}  & \multirow{3}{*}{MSA $\rightarrow$ MSA}  &     \textit{QALB 2014}  &\newcite{mohit2014first} &  $19.4$K     & $1$K & $968$  \\ 
                    &   &    \textit{QALB 2015}    &\newcite{rozovskaya2015second} &  $310$  & $154$  & $158$  \\  
                      &   &    \textit{ZAEBUC}    &\newcite{habash-palfreyman-2022-zaebuc} &  $27$K  & $3.3$K  & $3.3$K  \\\midrule

\multirow{3}{*}{\textbf{Paraphrase}}  &
\multirow{3}{*}{MSA $\rightarrow$ MSA}  & 

% \textit{AraPara}  &\newcite{nagoudi2022_arat5} &  $116.4$K     & $6.1$K & $-$  \\ &  & 
                        \textit{ASEP}  & \newcite{cer2017semeval}   & $116.4$K  & $6.1$K  & $600$ \\ 
                    &  &     \textit{APB} &  \newcite{alian2019towards} &  $808$     & $202$ & $101$  \\ 
                    &  &    \textit{TaPaCo}  & \newcite{scherrer-2020-tapaco}   & $2.1$K  & $299$  & $605$ \\ \midrule

% \toprule
\end{tabular}
}
\caption{Statistics of our\textit{~\dolphin~} benchmark across the different task clusters. For the QA task, we use the  Arabic machine translated SQuAD (AR-XTREME\textsubscript{train}) from \newcite{pmlr-v119-hu20b} as Train for ARCD, MLQA, and XQuAD. We also use AR-XTREME\textsubscript{dev} as Dev for XQuAD and TyiQA, respectively. For ASEP~\cite{cer2017semeval} test set in the summarization task, we use AraPara\textsubscript{Train} and AraPara\textsubscript{Dev}.    }  
\label{tab:dolphin-stats}
\end{table*}

\section{Arabic and Multilingual S2S LLMs} \label{app:llms}

In this section, we list the Arabic and multilingual sequence-to-sequence (S2S) pretrained LMs we finetune on~\dolphin. 

%, including AraT5, AraBART, mT5, mBART, and mT0.
\noindent{\textbf{AraT5.}}~\cite{nagoudi2022_arat5} is an adaptation of the T5 model specifically designed for the Arabic language. It is pre-trained on a large ($248$GB of Arabic text) diverse  (MSA and Arabic dialects) dataset to effectively handle different Arabic tasks.  In addition to Arabic, AraT5's vocabulary covers $11$ other languages. In this work, we evaluate a new in-house version of AraT5 dubbed AraT5\textsubscript{v2}.

\noindent{\textbf{AraT5\textsubscript{v2}.}} Our analysis shows that AraT5 requires a large number of epochs to converge, making it an expensive model. For this reason, we pretrain a new version of the model from scratch exploiting a larger ($\sim400$GB) and more diverse pretraining dataset than used by~\cite{nagoudi2022_arat5}. As we show in our results, the new model converges faster than AraT5 and achieves better results under our cap of $20$ epochs for finetuning across all models.

\noindent{\textbf{AraBART.}} \cite{eddine2022arabart} is a model based on the encoder-decoder BART base architecture~\cite{lewis2019bart}, featuring six encoder and $6$ decoder layers. It is pretrained on the same corpus as AraBERT~\cite{antoun2020arabert}, with reversed preprocessing for more natural text generation. AraBART is designed for various NLP tasks, demonstrating robust performance across different tasks in the Arabic language. 

% \subsection{Multilingual S2S LMs}

\noindent\textbf{mBART.} A multilingual encoder-decoder model proposed by ~\newcite{liu2020multilingual}. mBART is pretrained by denoising full texts in $50$ languages, including Arabic. Then, it is finetuned on parallel MT data contains a total of $230$M parallel sentences under three settings: individually toward English and vice versa (i.e., \textit{many-to-English}, and \textit{English-to-many}), or between multiple languages simultaneously (many-to-many). 

\noindent\textbf{mT5.}~\cite{xue2020mt5} is a multilingual variant of the of \textbf{T}ext-\textbf{t}o-\textbf{T}ext \textbf{T}ransfer \textbf{T}ransformer model (T5)~\cite{raffel2019exploring} that covers 101 languages. It is pretrained on a new Common Crawl-based dataset ($\sim26.76$TB), designed to achieve SOTA performance on a variety of multilingual NLP tasks such as question answering, document summarization, and MT.

\noindent{\textbf{mT0.}}~\cite{muennighoff2022crosslingual} is a group of sequence-to-sequence  models ranging in size between $300$M to $13$B parameters trained to investigate the cross-lingual generalization through multitask finetuning.  The models are finetuned from preexisting mT5~\cite{xue2020mt5} multilingual language models using a cross-lingual task dataset called xP3. mT0 models can execute human instructions in many languages without any prior training.

% \section{\dolphin: Examples from Each Task}
% \label{app:examples}

% \input{Tables/examples}

\section{Leaderboard}\label{app:leaderboard}
%%%%%%%%%%%%%%%%%%%%%%%%%%%
%%%%%%%%%%%%%%%%%%%%%%%%%%%

\end{document}